\documentclass{article}

     \usepackage[preprint,nonatbib]{neurips_2024}

\usepackage{hyperref}

\usepackage{graphicx}
\usepackage{acronym}
\usepackage{amsmath}
\usepackage{amssymb}
\usepackage{xcolor}
\usepackage{booktabs}
\usepackage{placeins}
\usepackage{algorithm}
\usepackage{algpseudocode}
\usepackage{caption}
\usepackage{subcaption}
\usepackage{comment}
\usepackage[toc,page]{appendix}
\usepackage{enumitem}

\newcommand{\noteValue}[1]{{\color{gray} {\scriptsize [#1]}}}
\def\checkmark{\tikz\fill[scale=0.4](0,.35) -- (.25,0) -- (1,.7) -- (.25,.15) -- cycle;}

\title{UNCOVER: Unknown Class Object Detection for Autonomous Vehicles in Real-time}

\author{%
Lars Schmarje $\dagger,\ddagger$, Kaspar Sakmann $\dagger$, Reinhard Koch $\ddagger$, Dan Zhang $\dagger$\\
  $\dagger$Bosch Center for Artificial Intelligence\\
  $\ddagger$ Kiel University\\
  \texttt{science@schmarje-sh.de} \\
}

\begin{document}
\graphicspath{{images/}}

\newcommand{\pic}[3]{
	\begin{figure}[tb] 
		\centering
		\includegraphics[width=#3\linewidth]{#1}
		\caption{#2}
		\label{fig:#1}
	\end{figure}
}

\newcommand{\picLarge}[3]{
	\begin{figure*}[tb] 
		\centering
		\includegraphics[width=#3\linewidth]{#1}
		\caption{#2}
		\label{fig:#1}
	\end{figure*}
}

\newcommand{\picTwo}[4]{
	\begin{figure}[tb]
		\centering
		\begin{subfigure}[c]{0.48\linewidth}
			\centering		
			\includegraphics[width=0.95\linewidth]{#1-0}
			\subcaption{#2}	
			\label{fig:#1-0}	
		\end{subfigure}
		\begin{subfigure}[c]{0.48\linewidth}	
			\centering	
			\includegraphics[width=0.95\linewidth]{#1-1}
			\subcaption{#3}
			\label{fig:#1-1}	
		\end{subfigure}
		\caption{#4}
		\label{fig:#1}
	\end{figure}
}
\newcommand{\picTwoLegend}[4]{
	\begin{figure}[tb]
		\centering
  \begin{subfigure}[c]{0.98\linewidth}
			\centering		
			\includegraphics[width=0.95\linewidth]{#1-legend}
			\label{fig:#1-legend}	
		\end{subfigure}
		\begin{subfigure}[c]{0.48\linewidth}
			\centering		
			\includegraphics[width=0.95\linewidth]{#1-0}
			\subcaption{#2}	
			\label{fig:#1-0}	
		\end{subfigure}
		\begin{subfigure}[c]{0.48\linewidth}	
			\centering	
			\includegraphics[width=0.95\linewidth]{#1-1}
			\subcaption{#3}
			\label{fig:#1-1}	
		\end{subfigure}
		\caption{#4}
		\label{fig:#1}
	\end{figure}
}

\newcommand{\picThree}[5]{
	\begin{figure*}[tb]
		\centering
		\begin{subfigure}[c]{0.3\textwidth}
			\centering		
			\includegraphics[width=0.95\textwidth]{#1-0}
			\subcaption{#2}	
				\label{fig:#1-0}	
		\end{subfigure}
		\begin{subfigure}[c]{0.3\textwidth}	
			\centering	
			\includegraphics[width=0.95\textwidth]{#1-1}
			\subcaption{#3}	
				\label{fig:#1-1}	
		\end{subfigure}
		\begin{subfigure}[c]{0.3\textwidth}	
			\centering	
			\includegraphics[width=0.95\textwidth]{#1-2}
			\subcaption{#4}	
				\label{fig:#1-2}	
		\end{subfigure}
	
		\caption{#5}
		\label{fig:#1}
	\end{figure*}
}

\newcommand{\picFour}[6]{
	\begin{figure}[tb]
		\centering
		\begin{subfigure}[c]{0.24\textwidth}
			\centering		
			\includegraphics[width=0.9\textwidth]{#1-0}
			\subcaption{#2}	
			\label{fig:#1-0}	
		\end{subfigure}
		\begin{subfigure}[c]{0.24\textwidth}	
			\centering	
			\includegraphics[width=0.9\textwidth]{#1-1}
			\subcaption{#3}
			\label{fig:#1-1}	
		\end{subfigure}
  \begin{subfigure}[c]{0.24\textwidth}
			\centering		
			\includegraphics[width=0.9\textwidth]{#1-2}
			\subcaption{#4}	
			\label{fig:#1-2}	
		\end{subfigure}
		\begin{subfigure}[c]{0.24\textwidth}	
			\centering	
			\includegraphics[width=0.9\textwidth]{#1-3}
			\subcaption{#5}
			\label{fig:#1-3}	
		\end{subfigure}
		\caption{#6}
		\label{fig:#1}
	\end{figure}
}

\newcommand{\picSix}[5]{
	\begin{figure}[tb]
		\centering
		\begin{subfigure}[c]{0.31\textwidth}
			\centering		
			\includegraphics[width=0.9\textwidth]{#1-0}
			\label{fig:#1-0}	
		\end{subfigure}
		\begin{subfigure}[c]{0.31\textwidth}	
			\centering	
			\includegraphics[width=0.9\textwidth]{#1-1}
			\label{fig:#1-1}	
		\end{subfigure}
  \begin{subfigure}[c]{0.31\textwidth}	
			\centering	
			\includegraphics[width=0.9\textwidth]{#1-2}
			\label{fig:#1-2}	
		\end{subfigure}
  \begin{subfigure}[c]{0.31\textwidth}
			\centering		
			\includegraphics[width=0.9\textwidth]{#1-3}
			\subcaption{#2}	
			\label{fig:#1-3}	
		\end{subfigure}
		\begin{subfigure}[c]{0.31\textwidth}	
			\centering	
			\includegraphics[width=0.9\textwidth]{#1-4}
			\subcaption{#3}
			\label{fig:#1-4}	
		\end{subfigure}
  \begin{subfigure}[c]{0.31\textwidth}	
			\centering	
			\includegraphics[width=0.9\textwidth]{#1-5}
			\subcaption{#4}
			\label{fig:#1-5}	
		\end{subfigure}
		\caption{#5}
		\label{fig:#1}
	\end{figure}
}

\newcommand{\picSixPreds}[2]{
	\begin{figure}[tb]
		\centering
		\begin{subfigure}[c]{0.48\textwidth}
			\centering		
			\includegraphics[width=0.95\textwidth]{#1-0}
			\subcaption{Original}	
			\label{fig:#1-0}	
		\end{subfigure}
		\begin{subfigure}[c]{0.48\textwidth}	
			\centering	
			\includegraphics[width=0.95\textwidth]{#1-1}
			\subcaption{GT}
			\label{fig:#1-1}	
		\end{subfigure}
  \begin{subfigure}[c]{0.48\textwidth}	
			\centering	
			\includegraphics[width=0.95\textwidth]{#1-2}
			\subcaption{UNCOVER}
			\label{fig:#1-2}	
		\end{subfigure}
  \begin{subfigure}[c]{0.48\textwidth}
			\centering		
			\includegraphics[width=0.95\textwidth]{#1-3}
			\subcaption{YOLOWorld}	
			\label{fig:#1-3}	
		\end{subfigure}
		\begin{subfigure}[c]{0.48\textwidth}	
			\centering	
			\includegraphics[width=0.95\textwidth]{#1-4}
			\subcaption{GDINO}
			\label{fig:#1-4}	
		\end{subfigure}
  \begin{subfigure}[c]{0.48\textwidth}	
			\centering	
			\includegraphics[width=0.95\textwidth]{#1-5}
			\subcaption{RPL}
			\label{fig:#1-5}	
		\end{subfigure}
		\caption{#2}
		\label{fig:#1}
	\end{figure}
}

\newcommand{\tblIssues}{	
\setlength{\tabcolsep}{12pt}
\begin{table*}

		\centering

  		\caption{
Evaluation of algorithms for detecting safety-critical unknown objects --
"Known Detection" indicates the ability to identify training dataset objects.
"Unknown Detection" shows capability to recognize new, unseen objects.
"Control Detection" evaluates the control over identifying unknown objects, crucial for nuanced responses.
"Real-Time" marks the suitability for time-sensitive applications like autonomous driving.
}
		 \label{tbl:issues} 
		
		\resizebox{\linewidth}{!}{%
		\begin{tabular}{l   c  c    c c    c }
			\toprule
   Algorithm (Family) & Examples & Known Detection & Unknown Detection & Control Detection  & Real-Time \\
   \midrule
Object Detection & \cite{Ge2021yolox,Tian2019fcos} & \checkmark & & & \checkmark\\
Open World Object Detection & \cite{Zohar2023prob,Gupta2022owdetr} & \checkmark & \checkmark & & \\
Anomaly Segementation & \cite{Tian2022pebal,Liu2023rpl} & \checkmark & \checkmark & \checkmark &  \\
Open Vocabulary & \cite{Liu2023GDino,Cheng2024YoloWorld} & \checkmark & \checkmark  & \checkmark  & ? \\
UNCOVER (Ours) & &  \checkmark & \checkmark & \checkmark & \checkmark \\

			\bottomrule
		\end{tabular}
		}

\end{table*}
\setlength{\tabcolsep}{6pt}
}

\newcommand{\tblResultsanoseg}{	
\begin{table*}[tb]

		\centering
            
  		\caption{
   Comparison of the anomaly segmentation methods PEBAL and RPL. All methods are trained on Cityscapes~\cite{cityscapes} where MS COCO~\cite{mscoco} are exploited as OOD data. Best number is bold and 2nd best within 50\% margin is italic. %
		    }
		 \label{tbl:Results_anoseg} 
		
		\resizebox{0.9\linewidth}{!}{%
		\begin{tabular}{l c  c  c    c c   c c    c}
			\toprule
			 &   Runtime  & \multicolumn{2}{c}{Cityscapes}   & \multicolumn{1}{c}{FS L\&F} & \multicolumn{1}{c}{Anomaly} &  \multicolumn{1}{c}{Obstacle} \\ 			
		 \cmidrule(r){2-2} \cmidrule(r){3-4} \cmidrule(r){6-6} \cmidrule(r){7-7} \cmidrule(r){8-8}  \cmidrule(r){5-5} %
  Method  &  FPS $\uparrow$ %
  & mAP $\uparrow$ & R@100 $\uparrow$  &    R@100 $\uparrow$ & R@100 $\uparrow$ &  R@100 $\uparrow$  \\
\midrule
PEBAL~\cite{Tian2022pebal} &2.80%
&  N/A & 10.54    & 44.75  & 12.50  & 53.33\\
RPL~\cite{Liu2023rpl}   & 3.26 %
&  N/A &  \textbf{20.80}     &\textbf{ 70.17}& 18.75  & \textbf{84.44}  \\
\midrule
UNCOVER &  \textbf{26.29} %
& \textbf{35.94}  & \textit{15.62}  &\textit{ 49.72 } & \textbf{100.00} & \textit{82.22 }  \\ 

		\bottomrule
		\end{tabular}
		}

\end{table*}
}

\newcommand{\tblResultsopenvoc}{	
\begin{table*}[tb]

		\centering
            
  		\caption{Comparison with the open-vocabulary object detection models. Both GDINO~\cite{Zhao2024mmgdino} and YOLOWorld~\cite{Cheng2024YoloWorld} are trained on large-scale object detection data and demonstrate strong zero-shot generalization capabilities cross benchmarks. Our model is respectively trained with Cityscapes~\cite{cityscapes} and BDD100k~\cite{Yu2020BDD} together with LVIS~\cite{Gupta2019Lvis} as the OOD data.
    Best is marked bold and 2nd best is italic.
		    }
		 \label{tbl:Results_openvoc} 
		
		\resizebox{\linewidth}{!}{%
		\begin{tabular}{l c  c  c    c c    c c     c c      c c        c}
			\toprule
			     & Runtime  & \multicolumn{2}{c}{Cityscapes} & \multicolumn{2}{c}{BDD100K}  & \multicolumn{1}{c}{FS L\&F} & \multicolumn{1}{c}{Anomaly} &  \multicolumn{1}{c}{Obstacle} \\ 			
		 \cmidrule(r){3-4} 	\cmidrule(r){5-6}  \cmidrule(r){7-7} \cmidrule(r){8-8}  \cmidrule(r){9-9} \cmidrule(r){2-2}
 Method  &   FPS $\uparrow$ %
  & mAP $\uparrow$ & R@100 $\uparrow$  &  mAP  $\uparrow$ & R@100 $\uparrow$ &  R@100 $\uparrow$ & R@100 $\uparrow$ &  R@100 $\uparrow$  \\
\midrule
GDINO~\cite{Zhao2024mmgdino} &  6.05 %
& 32.76  & \textit{30.54}  & \textit{24.58}  & \textit{78.41}  & \textbf{74.03}  & \textbf{100.00} & \textbf{93.33} \\  
YOLOWorld~\cite{Cheng2024YoloWorld} &  \textbf{30.59} %
& 22.14  &  12.23  & 21.44 &  29.24 & 33.70  &  \textbf{100.00}  &   62.22  \\ 
\midrule
UNCOVER (CS) &  \textit{26.29 } %
& \textbf{35.91}  & 15.71  & 14.11  & 39.42  & \textit{58.56}  & \textit{93.75}  & 77.78   \\ 
UNCOVER (BDD) & \textit{26.29} %
& 25.14  & 17.86  & \textbf{30.33}  & \textbf{84.35}  & 55.80  & \textbf{100.00} & \textit{88.89}   \\ 

			\bottomrule
		\end{tabular}
		}

\end{table*}
}

\newcommand{\tblResultsocc}{	
\begin{table*}[tb]

		\centering
            
  		\caption{
   Comparison of objectness measures for open-world object detection using UNCOVER trained on CS~\cite{cityscapes} with OOD data from  LVIS~\cite{Gupta2019Lvis}.
    Best results are in bold.
		    }
		 \label{tbl:results_occ} 
		
	\resizebox{\linewidth}{!}{%
		\begin{tabular}{l  c    c c    c c     c c      c c        c}
			\toprule
			  & \multicolumn{2}{c}{Cityscapes} & \multicolumn{2}{c}{BDD100K}  & \multicolumn{1}{c}{FS L\&F} & \multicolumn{1}{c}{Anomaly} &  \multicolumn{1}{c}{Obstacle} \\ 			
		 \cmidrule(r){2-3} 	\cmidrule(r){4-5}  \cmidrule(r){7-7} \cmidrule(r){6-6}  \cmidrule(r){8-8}
Objectness measure  & mAP $\uparrow$ &
  R@100 $\uparrow$ & mAP $\uparrow$ & R@100 $\uparrow$ &  R@100 $\uparrow$ & R@100 $\uparrow$ &  R@100 $\uparrow$  \\
  \midrule
\emph{Obj.} score &  34.42 &13.21 & 12.40 & 39.38 & 48.62 & \textbf{93.75} & 75.56 \\ 
IoU score & 35.83 &11.70 & 11.51  & 39.32 & 52.49 & \textbf{93.75} & 73.33 \\
Occ. score (Ours) & \textbf{35.91 }&\textbf{15.71} &\textbf{14.11} & \textbf{39.42} & \textbf{58.56} & \textbf{93.75} & \textbf{77.78} \\
			\bottomrule
		\end{tabular}
	}

\end{table*}
}

\newcommand{\tblResultsBroad}{	
\begin{table*}

		\centering

  		\caption{Broader Results, DBSE applied for Cityscape and Fishyscapes, $^\dagger$ converted from task which considers Traffic Light/sign as known class
		    }
		 \label{tbl:Resultsv2} 
		
		\resizebox{\linewidth}{!}{%
		\begin{tabular}{l   c  c    c c   c c c        c}
			\toprule
			Dataset & \multicolumn{1}{c}{Cityscapes} & \multicolumn{1}{c}{FS L\&F}  & \multicolumn{1}{c}{BDD100K}  & \multicolumn{1}{c}{Anomaly} &  \multicolumn{1}{c}{Obstacle} & Runtime\\ 
			
			\cmidrule(r){2-3} \cmidrule(r){4-5}  \cmidrule(r){7-7}  \cmidrule(r){8-8}  \cmidrule(r){6-6}
  Method          &   R@100 $\uparrow$  &  R@100 $\uparrow$   & R@100 $\uparrow$ & R@100 $\uparrow$ & R@100 $\uparrow$ &  FPS $\uparrow$  \\
\midrule

PEBAL        & 10.54  & 44.75  &  N/A$^\dagger$  & 12.50  & 53.33 & 2.80\\
RPL         &  20.80   & 70.17& N/A$^\dagger$  & 18.75  & 84.44   & 3.26 \\
\midrule

YOLOWorld          & 12.23   & 33.70 &  29.24  & 100.00 & 62.22 & 30.59\\
GDino        & 30.54  & 74.03& 78.41  & 100.00 & 93.33 & 6.05  \\

\midrule
Ours (CS)       & 15.71  & 57.46 &  39.42  & 93.75  & 77.78 & 26.29 \\
Ours (BDD)          & 13.93  &  55.25 & 84.85  & 100.00 & 88.89 & 26.29 \\

			\bottomrule
		\end{tabular}
		}

\end{table*}
}

\newcommand{\tblow}{	
\begin{table*}

		\centering

  		\caption{Comparison with the open world object detection models, i.e., PROB~\cite{Zohar2023prob} and OW-DETR~\cite{Gupta2022owdetr}, with our proposed UNCOVER. All of them are trained on Cityscapes~\cite{cityscapes}.
		    }
		 \label{tbl:ow} 
		
		\resizebox{\linewidth}{!}{%
		\begin{tabular}{l  c  c  c    c c c    c c     c c      c c        c}
			\toprule
			Dataset & \multicolumn{2}{c}{Cityscapes} & \multicolumn{2}{c}{BDD100K}  & \multicolumn{1}{c}{FS L\&F} & \multicolumn{1}{c}{Anomaly} &  \multicolumn{1}{c}{Obstacle} & Runtime\\ 
			
			\cmidrule(r){2-3} \cmidrule(r){4-5}  \cmidrule(r){6-6}  \cmidrule(r){8-8}  \cmidrule(r){7-7}
  Method          & mAP $\uparrow$ &  R@100 $\uparrow$ &  mAP $\uparrow$ &  R@100 $\uparrow$ &    R@100 $\uparrow$ & R@100 $\uparrow$ &  R@100 $\uparrow$ &  FPS $\uparrow$  \\
\midrule

PROB                                              & 23.06  & 10.54  &  7.36  &   0.57  &  4.42  &  56.25  &  6.67 & 14.81 \\ 
OW-DETR                                           & 18.67  &  4.91  &  6.54  &   0.48  &    4.97  &  25.00  &   6.67  & 15.27 \\

\midrule

UNCOVER (CS)   %
&35.91  & 15.71  & 14.11  & 39.42  & 58.56  & 93.75  & 77.78 & 26.29    \\

			\bottomrule
		\end{tabular}
		}

\end{table*}
}

\newcommand{\tblFull}{	
\begin{table*}

		\centering

  		\caption{More evaluation results with DFR
		    }
		 \label{tbl:full} 
		
		\resizebox{\linewidth}{!}{%
		\begin{tabular}{l  c  c  c    c c c    c c     c c      c c        c}
			\toprule
			Dataset & \multicolumn{3}{c}{Cityscapes} & \multicolumn{2}{c}{BDD100K}  & \multicolumn{3}{c}{FS L\&F} & \multicolumn{2}{c}{Anomaly} &  \multicolumn{2}{c}{Obstacle} & Runtime\\ 
			
			\cmidrule(r){2-4} \cmidrule(r){5-7}  \cmidrule(r){12-13}  \cmidrule(r){8-9}  \cmidrule(r){10-11}
  Method          & mAP $\uparrow$ &  FPR@100  $\downarrow$ & R@100 $\uparrow$ &  mAP $\uparrow$ &  FPR@100  $\downarrow$ & R@100 $\uparrow$ &    FPR@100  $\downarrow$ & R@100 $\uparrow$ &FPR@100  $\downarrow$ & R@100 $\uparrow$ &  FPR@100  $\downarrow$ & R@100 $\uparrow$ &  FPS $\uparrow$  \\
\midrule

PROB                                              & 23.06  &  4.5  & 10.54  &  7.36  &  0.3  &  0.57  &  4.3  &  4.42  &  0.0  & 56.25  &  0.0  &  6.67 & 14.81 \\ 
OW-DETR                                           & 18.67  &  3.5  &  4.91  &  6.54  &  1.3  &  0.48  &  3.8  &  4.97  &  0.0  & 25.00  &  0.4  &  6.67  & 15.27 \\
\midrule 
PEBAL                                             &  0.00  &  3.9  & 11.16  &  0.00  &  0.0  &  0.08  &  2.0  & 45.86  &  0.0  & 12.50  &  0.0  & 53.33  & 2.80 \\ 
RPL                                               &  0.00  &  6.5  & 20.80  &  0.00  &  0.1  &  0.24  &  7.6  & 69.06  &  0.0  & 18.75  &  0.0  & 84.44 & 3.26 \\      
\midrule
YOLOWorld                                        & 21.69  &  9.1  & 13.30  & 21.44  &  1.0  & 29.24  &  46.0  & 35.36  &  0.00  & 100.00 &  0.0  & 62.22& 20.39  \\ 

GDino                                             & 32.76  &  1.9  & 30.54  & 24.58  &  0.0  & 78.41  &  5.4  & 74.03  & 172.3  & 100.00 &  0.0  & 93.33 & 6.05 \\ 
\midrule
YOLOX + UNCOVER (CS)                              & 35.91  &  4.6  & 15.71  & 14.11  &  0.0  & 39.42  &  8.3  & 58.56  & 222.6  & 93.75  &  0.0  & 77.78  & 26.29 \\
YOLOX + UNCOVER (CS+OOD)                          & 35.96  &  12.8  & 35.36  & 14.87  &  0.0  & 36.14  &  9.9  & 53.04  & 222.1  & 100.00 &  0.0  & 80.00  & 26.29 \\
YOLOX + UNCOVER (BDD)                             & 25.14  &  2.5  & 17.86  & 33.05  &  0.0  & 84.85  &  2.8  & 55.80  & 261.6  & 100.00 &  0.0  & 88.89  & 26.29 \\

			\bottomrule
		\end{tabular}
		}

\end{table*}
}

\newcommand{\tblPost}{	
\begin{table*}[tb]

		\centering

  		\caption{
    Results with added occupancy post-processing --
    The scores between without and with occupancy are given.
    Improvements over over 20\% for FPR@100 and 1\% for R@100 are marked bold.
		    }
		 \label{tbl:post} 
		
		\resizebox{\linewidth}{!}{%
		\begin{tabular}{l l  l l l  l l l l l}
			\toprule
			Dataset & \multicolumn{3}{c}{Cityscapes} & \multicolumn{2}{c}{FS L\&F} & \multicolumn{2}{c}{Anomaly}  & \multicolumn{2}{c}{Obstacle}   \\ 
			
			\cmidrule(r){2-4} \cmidrule(r){5-6}  \cmidrule(r){7-8}  \cmidrule(r){9-10}  
  Method      & mAP  $\uparrow$    &  R@100 $\uparrow$  &  FPR@100  $\downarrow$ &  R@100 $\uparrow$ & FPR@100  $\downarrow$ & R@100 $\uparrow$  &FPR@100  $\downarrow$ & R@100 $\uparrow$ & FPR@100  $\downarrow$  \\
\midrule

Ours & 35.91 & 15.71 & 4.6 & 58.56 & 8.3 & 93.75 & 22.2 & 77.78 & 10.0 \\

\midrule 
		
PEBAL      &  0.0 & 11.16 & 3.9 & 45.86 & 2.0 & 62.50 & 68.1 & 53.33 & 3.1 \\
+ Occupancy      &  35.90 (+\%) & 11.16 (+0.0\%)&  2.6 (-\%) & 45.86 (+\%) & 1.6 (-\%) & 62.50 (+ 0.0\%)  & 45.7 (-\%) & 53.33 (+0.00\%)  & 2.7 (-\%)\\
RPL        & 0.0 & 20.80 & 6.5 & 69.06 & 7.6 & 81.25 & 252.4 & 84.44 & 2.3 \\
+ Occupancy      & 35.90 (+\%) & 20.71 (+\%) & 4.3 (-\%) & 69.06 (+\%) & 5.7 (-\%) & 100.00 (+0.00\%) & 199.9 (-\%) & 84.44 (+0.0\%) & 1.7 (-\%)\\

\midrule
YoloWorld & 21.69 & 13.30 & 9.1 &  35.36 & 46.0 & 100.00 & 543.7 & 62.22 & 383.3\\
+ Occupancy &   34.94 (+\%) & 20.71 (+\%) & 2.8 (-\%) & 51.38 (+\%) & 10.2 (-\%) & 100.0 (+ 0.0\%) & 199.9 (-\%) & 77.78 (+\%) & 13.6 (-\%) \\

GDino     & 32.76 & 30.54 & 1.9 & 74.03 & 5.4 & 100.00 & 172.3 & 100.00 & 20.5 \\
+ Occupancy     & 35.38 (+\%) & 30.00 (+\%) & 1.0 (-\%) & 72.93 (+\%) & 3.6 (-\%) &  100.00 (+ 0.0\%) & 79.2 (-\%) & 100.00  (+0.00\%) & 9.4 (-\%)\\

			\bottomrule
		\end{tabular}
	}

\end{table*}
}

\newcommand{\tblFullDBSE}{	
\begin{table*}[tb]

		\centering

  		\caption{
    Results for Depth Based False Positive Reduction (DFR) --
    The relative changes between without and with DFR are given as well as the score with DFR in brackets.
    Improvements over over 20\% for FPR@100 and 1\% for R@100 are marked bold.
		    }
		 \label{tbl:dbseFull} 
		
		\begin{tabular}{l   l l l  l }
			\toprule
			Dataset & \multicolumn{2}{c}{Cityscapes} & \multicolumn{2}{c}{FS L\&F}  \\ 
			
			\cmidrule(r){2-3} \cmidrule(r){4-5} 
  Method          &  FPR@100  $\downarrow$ & R@100 $\uparrow$  &  FPR@100  $\downarrow$ & R@100 $\uparrow$  \\
\midrule

PEBAL      &  +0.0\% \noteValue{3.9} &\textbf{+5.9\%} \noteValue{11.16} & \textbf{-28.6\%} \noteValue{2.0} & \textbf{+2.5\%} \noteValue{45.86}\\
RPL        &-8.5\% \noteValue{6.5} & +0.0\%  \noteValue{20.80} & \textbf{-20.8\%} \noteValue{7.6} & -1.6\% \noteValue{69.06}  \\
\midrule
YOLOWorld &  +3.4\% \noteValue{9.1} &\textbf{+8.7\%} \noteValue{13.30} &  \textbf{-24.3\%}\noteValue{46.0} &\textbf{+4.9\%} \noteValue{35.36}\\
GDino     & \textbf{-24.0\%} \noteValue{1.9} & +0.0\% \noteValue{30.54}  &  \textbf{-38.6\%} \noteValue{5.4} & +0.0\% \noteValue{74.03}  \\

\midrule
Ours (CS)  &\textbf{-23.3\%} \noteValue{4.6} & +0.0\% \noteValue{15.71} &  \textbf{-25.2\%} \noteValue{8.3}  & +0.0 \noteValue{58.56}  \\
Ours (BDD) & +13.6\% \noteValue{2.5} & \textbf{+28.2\%}\noteValue{17.86} & \textbf{-44.0\%} \noteValue{2.8} & \textbf{+1.0\%} \noteValue{55.80} \\

			\bottomrule
		\end{tabular}

\end{table*}
}

\newcommand{\tblDBSE}{	
\begin{table*}[tb]

		\centering

  		\caption{Results for Depth Based False Positive Reduction (DFR) --
    Relative changes between without and with DFR are given in brackets. Improvements over over 20\% for FPR@100 and 1\% for R@100 are marked bold.
		    }
		 \label{tbl:dbse} 
		
		\resizebox{0.9\linewidth}{!}{%
		\begin{tabular}{l  l l l l  }
			\toprule
			Dataset & \multicolumn{2}{c}{Cityscapes} & \multicolumn{2}{c}{FS L\&F}  \\ 
			
			\cmidrule(r){2-3} \cmidrule(r){4-5} 
  Method          &  FPR@100  $\downarrow$ & R@100 $\uparrow$  &  FPR@100  $\downarrow$ & R@100 $\uparrow$  \\
\midrule

PEBAL      & 3.9 & 10.54 & 2.8 & 44.75\\
+ DFR      &  3.9 (+0.0\%) & 11.16 (\textbf{+5.9\%})&  2.0 (\textbf{ -28.6\%}) &  45.86 (\textbf{+2.5\%})\\
RPL        &7.1 & 20.80 & 9.6 & 70.17\\
+ DFR        & 6.5 (-8.5\%) &  20.80 (+0.0\%) & 7.6 (\textbf{-20.8\%}) &  69.06 (-1.6\%)  \\
\midrule
YOLOWorld & 8.8 & 12.23 & 60.8 &  33.70 \\
+ DFR &   9.1 (+3.4\%) &13.30 (\textbf{+8.7\% })&  46.0 (\textbf{-24.3\%}) & 35.36 (\textbf{+4.9\%})\\
GDINO     & 2.5 & 30.54 & 8.8 & 74.03\\
+ DFR     & 1.9 (\textbf{-24.0\%}) &  30.54 (+0.0\%)  &   5.4 (\textbf{-38.6\%}) &  74.03 (+0.0\%)  \\

\midrule
UNCOVER (CS)  & 6.0 & 15.71 & 11.1 & 58.56\\
+ DFR  & 4.6 (\textbf{-23.3\%}) &  15.71 (+0.0\%) &   8.3 (\textbf{-25.2\%}) &  58.56 (+0.0\%)  \\
UNCOVER (BDD) & 2.2 & 13.93 & 0.5 & 55.25\\
+ DFR & 2.5 (+13.6\%) & 17.86 (\textbf{+28.2\%}) & 2.8 (\textbf{-44.0\%})& 55.80 (\textbf{+1.0\%}) \\

			\bottomrule
		\end{tabular}
	}

\end{table*}
}

\newcommand{\tblAblations}{	
\begin{table}[tb]

		\centering

  		\caption{Ablation of the OOD classification and occupancy prediction for unknown object detection. Models are trained on Cityscapes~\cite{cityscapes} with LVIS~\cite{Gupta2019Lvis} as OOD data. The much reduced FP without OOD class is due to reduced OOD detections in general, as we can see from the number drops on Recall.}
		 \label{tbl:Ablations} 
		
		\resizebox{0.7\linewidth}{!}{%
		\begin{tabular}{l   c c  c    c c  }
			\toprule
	
			 & \multicolumn{3}{c}{Cityscapes} &  \multicolumn{2}{c}{FS L\&F}\\ 
			
			\cmidrule(r){2-4} \cmidrule(r){5-6}
    Method      & mAP$\uparrow$  & FPR@100 $\downarrow$ & R@100$\uparrow$  &  FPR@100 $\downarrow$ & R@100 $\uparrow$  \\
\midrule
YOLOX~\cite{Ge2021yolox} & 36.03 & - & - &- &- \\
\midrule
UNCOVER                                     & 35.91  &  \textbf{6.0} & \textbf{15.71}&  \textbf{12.6} & \textbf{58.56}  \\

\midrule

- OOD class.                                         & \textbf{36.11}  & \textcolor{red}{0.3}  & 13.57  &  \textcolor{red}{0.8}  & 48.07  \\

- Occ. pred.                                  & 35.92 & 21.3  &  8.39  & 20.9  & 55.80  \\

			\bottomrule
		\end{tabular}
	}

\end{table}

	}

 \newcommand{\tblAblationsFull}{	
\begin{table}

		\centering

  		\caption{Results
		    }
		 \label{tbl:Ablations} 
		
		\resizebox{\linewidth}{!}{%
		\begin{tabular}{l   c c  c    c c  }
			\toprule
	
			Dataset & \multicolumn{3}{c}{Cityscapes} &  \multicolumn{2}{c}{FS L\&F}\\ 
			
    Method        & mAP$\uparrow$  & FPR@100 $\downarrow$ & R@100$\uparrow$  &  FPR@100 $\downarrow$ & R@100 $\uparrow$  \\
\midrule
Ours                                              & 35.91  &  0.46  & 15.71  &  0.83  & 58.56  \\
Occ on cl.                                        & 35.68  &  0.25  & 14.73  &  0.73  & 51.93  \\
Decoupled Occ.                                    & 35.38  &  0.40  & 15.18  &  0.92  & 56.91  \\
Coco                                              & 35.94  &  0.22  & 15.62  &  0.68  & 49.72  \\
w/o ucDet                                         & 36.11  &  0.03  & 13.57  &  0.08  & 48.07  \\
w/o caDet                                         & 35.92  &  5.56  & 10.71  &  1.25  & 54.14  \\
w/o Novel Loss                                    & 34.75  &  0.66  & 12.86  &  0.77  & 50.83  \\
w/o DBSE                                          & 35.91  &  0.60  & 15.71  &  1.26  & 58.56  \\
w/o uDis                                          & 35.91  &  1.94  &  9.02  &  1.80  & 56.91  \\
w/o caDet & uDis                                  & 35.92  &  2.13  &  8.39  &  2.09  & 55.80  \\
			\bottomrule
		\end{tabular}
		}

\end{table}

	}

 \newcommand{\tblDepthsCS}{	
\begin{table}

		\centering

  		\caption{Results for \autoref{fig:dbse_table-0} for reproducibility 
		    }
		 \label{tbl:DepthCS} 
		
		\resizebox{\linewidth}{!}{%
		\begin{tabular}{l   c c  c  c c    c c c  c c   }
			\toprule
	
			Dataset & \multicolumn{4}{c}{Cityscapes} &  \multicolumn{4}{c}{FS L\&F} \\ 
   \cmidrule(r){2-5} \cmidrule(r){6-9} 
			 Depth & \multicolumn{2}{c}{Stereo} &  \multicolumn{2}{c}{Monocular} & \multicolumn{2}{c}{Stereo} &  \multicolumn{2}{c}{Monocular} \\
			
			\cmidrule(r){2-3} \cmidrule(r){4-5}  \cmidrule(r){6-7}  \cmidrule(r){8-9}

       method                                       &    FPR@100 & R@100  &  FPR@100 & R@100  &   FPR@100 & R@100  &  FPR@100 & R@100   \\
                                                 \midrule
$\mu = 0$                                       &  0.60  & 15.71  &  0.60  & 15.71  &  1.26  & 58.56  &  1.26  & 58.56  \\
$\mu = 0.1$                                     &  0.61  & 15.98  &  0.60  & 15.71  &  1.27  & 55.25  &  1.26  & 58.56  \\
$\mu = 0.2$                                     &  0.62  & 15.80  &  0.59  & 15.71  &  1.26  & 52.49  &  1.25  & 58.56  \\
$\mu = 0.25$                                    &  0.63  & 15.80  &  0.55  & 15.71  &  1.27  & 50.28  &  1.17  & 58.56  \\
$\mu = 0.3$                                     &  0.65  & 15.71  &  0.46  & 15.71  &  1.26  & 49.17  &  0.83  & 58.56  \\
$\mu = 0.35$                                    &  0.68  & 15.80  &  0.40  & 15.80  &  1.23  & 48.62  &  0.54  & 56.91  \\
$\mu = 0.4$                                     &  0.70  & 15.54  &  0.36  & 15.98  &  1.22  & 46.96  &  0.42  & 54.14  \\
$\mu = 0.42$                                    &  0.71  & 15.54  &  0.34  & 15.98  &  1.22  & 46.41  &  0.39  & 53.59  \\
$\mu = 0.44$                                    &  0.72  & 15.54  &  0.33  & 15.98  &  1.20  & 45.86  &  0.35  & 53.04  \\
$\mu = 0.46$                                    &  0.74  & 15.36  &  0.32  & 15.98  &  1.18  & 45.30  &  0.31  & 51.93  \\
$\mu = 0.48$                                    &  0.75  & 15.45  &  0.30  & 16.07  &  1.21  & 44.75  &  0.30  & 50.83  \\
$\mu = 0.5$                                     &  0.78  & 15.27  &  0.28  & 16.07  &  1.20  & 43.65  &  0.26  & 48.07  \\
$\mu = 0.52$                                    &  0.80  & 15.09  &  0.26  & 16.16  &  1.21  & 39.23  &  0.23  & 45.86  \\
$\mu = 0.54$                                    &  0.82  & 14.91  &  0.24  & 16.16  &  1.24  & 38.12  &  0.22  & 44.75  \\
$\mu = 0.56$                                    &  0.85  & 14.82  &  0.22  & 15.98  &  1.23  & 36.46  &  0.18  & 43.09  \\
$\mu = 0.58$                                    &  0.88  & 14.73  &  0.20  & 15.98  &  1.25  & 35.36  &  0.15  & 39.78  \\
$\mu = 0.6$                                     &  0.93  & 14.37  &  0.19  & 15.89  &  1.25  & 34.25  &  0.12  & 36.46  \\
$\mu = 0.65$                                    &  1.11  & 14.02  &  0.17  & 15.80  &  1.32  & 27.62  &  0.06  & 32.04  \\
$\mu = 0.7$                                     &  1.52  & 11.79  &  0.13  & 15.71  &  1.40  & 20.99  &  0.04  & 24.86  \\
$\mu = 0.75$                                    &  2.21  &  9.55  &  0.12  & 15.45  &  1.41  & 12.71  &  0.02  & 20.44  \\

			\bottomrule
		\end{tabular}
		}

\end{table}

	}

 \newcommand{\tblDepthsBDD}{	
\begin{table}

		\centering

  		\caption{Results for \autoref{fig:dbse_table-1} for reproducibility 
		    }
		 \label{tbl:DepthBDD} 
		
		\resizebox{\linewidth}{!}{%
		\begin{tabular}{l   c c  c  c c    c c c  c c   }
			\toprule
	
			Dataset & \multicolumn{4}{c}{Cityscapes} &  \multicolumn{4}{c}{FS L\&F} \\ 
   \cmidrule(r){2-5} \cmidrule(r){6-9} 
			 Depth & \multicolumn{2}{c}{Stereo} &  \multicolumn{2}{c}{Monocular} & \multicolumn{2}{c}{Stereo} &  \multicolumn{2}{c}{Monocular} \\
			
			\cmidrule(r){2-3} \cmidrule(r){4-5}  \cmidrule(r){6-7}  \cmidrule(r){8-9}

       method                                       &    FPR@100 & R@100  &  FPR@100 & R@100  &   FPR@100 & R@100  &  FPR@100 & R@100   \\
                                                 \midrule
$\mu = 0$                                       &  0.29  & 17.86  &  0.29  & 17.86  &  0.44  & 55.80  &  0.44  & 55.80  \\
$\mu = 0.1$                                     &  0.29  & 17.86  &  0.29  & 17.86  &  0.41  & 53.04  &  0.44  & 55.80  \\
$\mu = 0.2$                                     &  0.30  & 17.86  &  0.29  & 17.86  &  0.41  & 51.93  &  0.42  & 55.80  \\
$\mu = 0.25$                                    &  0.30  & 17.86  &  0.28  & 17.86  &  0.41  & 49.17  &  0.39  & 55.80  \\
$\mu = 0.3$                                     &  0.30  & 18.04  &  0.25  & 17.86  &  0.40  & 48.07  &  0.28  & 55.80  \\
$\mu = 0.35$                                    &  0.31  & 17.77  &  0.23  & 17.86  &  0.40  & 47.51  &  0.19  & 55.25  \\
$\mu = 0.4$                                     &  0.33  & 17.68  &  0.20  & 17.95  &  0.40  & 46.96  &  0.16  & 54.70  \\
$\mu = 0.42$                                    &  0.34  & 17.59  &  0.19  & 17.95  &  0.39  & 44.20  &  0.16  & 53.59  \\
$\mu = 0.44$                                    &  0.35  & 17.68  &  0.17  & 17.95  &  0.40  & 43.65  &  0.13  & 51.93  \\
$\mu = 0.46$                                    &  0.36  & 17.68  &  0.16  & 18.04  &  0.39  & 43.09  &  0.11  & 50.83  \\   
$\mu = 0.48$                                    &  0.38  & 17.59  &  0.15  & 18.04  &  0.40  & 41.99  &  0.11  & 49.72  \\
$\mu = 0.5$                                     &  0.40  & 17.32  &  0.14  & 18.04  &  0.39  & 40.88  &  0.10  & 47.51  \\
$\mu = 0.52$                                    &  0.42  & 17.05  &  0.13  & 18.04  &  0.39  & 39.78  &  0.11  & 46.96  \\
$\mu = 0.54$                                    &  0.43  & 16.70  &  0.12  & 18.04  &  0.40  & 37.57  &  0.08  & 44.75  \\
$\mu = 0.56$                                    &  0.46  & 16.34  &  0.11  & 18.04  &  0.41  & 34.81  &  0.06  & 43.09  \\
$\mu = 0.58$                                    &  0.49  & 16.52  &  0.10  & 18.04  &  0.42  & 33.15  &  0.06  & 39.23  \\
$\mu = 0.6$                                     &  0.53  & 16.25  &  0.09  & 18.04  &  0.43  & 31.49  &  0.06  & 37.57  \\
$\mu = 0.65$                                    &  0.70  & 14.64  &  0.07  & 18.12  &  0.46  & 24.86  &  0.02  & 30.39  \\
$\mu = 0.7$                                     &  1.13  & 13.13  &  0.06  & 17.68  &  0.49  & 20.99  &  0.01  & 23.76  \\
$\mu = 0.75$                                    &  1.69  &  9.73  &  0.06  & 17.68  &  0.59  & 12.71  &  0.01  & 18.23  \\

			\bottomrule
		\end{tabular}
		}

\end{table}

	}

\maketitle

\begin{abstract}

Autonomous driving (AD) operates in open-world scenarios, where encountering unknown objects is inevitable. However, standard object detectors trained on a limited number of base classes tend to ignore any unknown objects, posing potential risks on the road. To address this, it is important to learn a generic rather than a class specific objectness from objects seen during training. %
We therefore introduce an occupancy prediction together with bounding box regression. It learns to score the objectness by calculating the ratio of the predicted area occupied by actual objects. To enhance its generalizability, we increase the object diversity by exploiting data from other domains via Mosaic and Mixup augmentation. The objects outside the AD training classes are classified as a newly added out-of-distribution (OOD) class. %
Our solution \textbf{UNCOVER}, for \textbf{UN}known \textbf{C}lass \textbf{O}bject detection for autonomous \textbf{VE}hicles in \textbf{R}eal-time, excels at achieving both real-time detection and high recall of unknown objects on challenging AD benchmarks. To further attain very low false positive rates, particularly for close objects, we introduce a post-hoc filtering step that utilizes geometric cues extracted from the depth map, typically available within the AD system. %

\end{abstract}

\section{Introduction}
Object detection is a core task in the perception stack of autonomous driving (AD) systems, in  which a model is trained to recognize objects from a pre-specified list of classes typical for AD, e.g., vulnerable road users, vehicles, traffic signs and traffic lights. %
However, achieving a high detection performance on these training classes alone is not sufficient; %
whenever an object hinders safe driving, a positive detection is required, even if it is from an unknown class, i.e., an out-of-distribution (OOD) object. It is particularly challenging to address such unknown object detection for AD.
Since autonomous vehicles often need to make swift decisions with limited on-device computation resources, the extra capability of unknown object detection must be acquired with a small complexity and inference latency increase~\cite{gao2021autonomous}. %
Moreover, improving the recall of unknown objects should not compromise the performance on the known classes, and even more importantly should not yield many false positive detections, as they can be equally dangerous in AD scenarios, e.g. for emergency brake systems. %

Current research on recognizing unknown objects in AD scenes focuses mainly on semantic segmentation~\cite{Tian2022pebal,Liu2023rpl,galesso2023cdnp,Rai2023Mask2Anomaly}, with well established benchmarks like Fishyscapes~\cite{Blum2021Fishy} and SegmentMeIfyouCan~\cite{Chan2021SMIYC}. Compared to semantic segmentation, object detection typically requires less inference-time complexity and can localize each instance individually. However, there is less focus on equipping real-time object detectors for AD with awareness of unknown objects. Beyond the AD domain, open-world object detection leverages unannotated objects in the training set, but real-world unknown objects might be completely novel. Another line of work identified that some objects outside the training classes may still be localized by the object detector, but misclassifying them as known classes deteriorates detection performance~\cite{Dhamija2020Elephant}.
Our work focuses on improving the recall of unknown objects, complementing these efforts.

\picThree{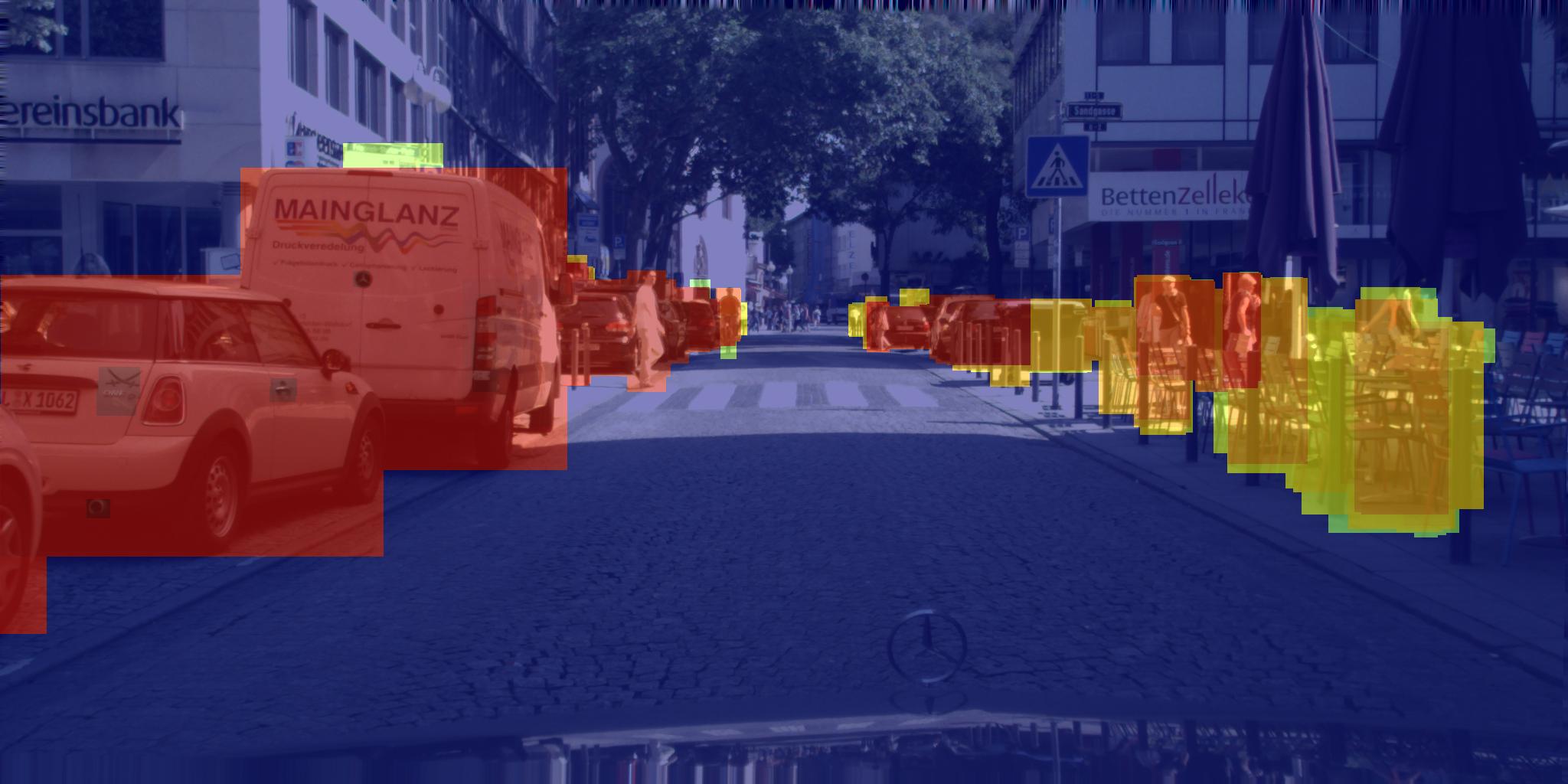}{Cityscapes~\cite{cityscapes}}{Fishyscapes~\cite{Blum2021Fishy}}{Anomaly~\cite{Chan2021SMIYC}}{Visualization of occupancy -- Here, UNCOVER was trained on Cityscapes~\cite{cityscapes}. Color code from blue, yellow, to red means low, medium and high occupancy. (a) UNCOVER predicts the highest occupancy on the known object classes from Cityscapes~\cite{cityscapes}, e.g, vehicle, person. (b) As OOD objects, the boxes on the ground from Fishyscapes~\cite{Blum2021Fishy} also have relatively high occupancy scores, compared to the background in blue. (c) It also responds to occluded objects, i.e., both vehicles and OOD object (trailer) from Anomaly~\cite{Chan2021SMIYC}. %
}

Aiming at real-time and low-complexity solutions, we hypothesize that the key to detecting unknown objects lies in learning a generic sense of objectness from the available training objects. %
In prior work on generic object detection~\cite{kim2021oln,huang2023good}, the Intersection-over-Union (IoU) score, initially proposed for quantifying bounding box quality, was leveraged as an objectness measure for detecting unknowns. In~\cite{Konan2022FCOSopen}, it was added to the one stage object detector FCOS~\cite{Tian2019fcos} for unknown-aware multi-class object detection. In this work, we propose a new objectness measure. Unlike the IoU score, the proposed occupancy score focuses less on bounding box quality, which typically needs supervision to be high. Instead, it focuses on evaluating whether the predicted area contains one or multiple objects, relaxing the localization quality constraint. We can observe from \autoref{fig:occupancy} that the occupancy score responds to objects from both the known and unknown classes, while it remains silent for \emph{stuff} classes (road, sky). 
As exposure to diverse objects enhances the acquisition of a more generic understanding of objectness, we enrich AD training with data from MS COCO~\cite{mscoco} and LVIS~\cite{Gupta2019Lvis}, using common augmentation techniques like Mosaic~\cite{bochkovskiy2020mosaic} and Mixup~\cite{zhang2018mixup}.
The former composes multiple images from different datasets into one image, providing exposure to OOD objects, while the latter interpolates the composed image with another AD image, helping to mitigate the domain gap between data domains.
For objects outside the AD training cases, we introduce an extra class, i.e., OOD class, in the classification head of the base object detector.

Overall, our architecture modification consists only of an extra class in the classification head and an occupancy prediction in the regression head, resulting in a small complexity increase. We call the resulting model UNCOVER. %
We further propose a post-hoc and optional filtering. Depth information, often available in AD systems, encodes geometric cues of objects, which complements the RGB-based appearance cues. Applying classic computer vision algorithms for depth change detection~\cite{sobel1968,fisher1996morph} helps to remove ghost object detection, such as drawings or shadows, as they don't have a geometric shape. This depth-based filtering is an interpretable algorithm helpful in reducing near-range false positive detections, thereby improving the safety of AD systems.
The architecture changes and the post-hoc filtering are modular by design to allow easier adoption but we show that the best results can be achieved with UNCOVER when combining them.

We extend our evaluation to include anomaly segmentation benchmarks~\cite{Blum2021Fishy,Chan2021SMIYC} by converting the respective mask annotations into bounding box formats for analysis
and assess false positives within specific regions on Cityscapes~\cite{cityscapes} and BDD100k~\cite{Yu2020BDD}.

In summary, this paper makes the following contributions for real-time AD:
\begin{itemize}[noitemsep,topsep=0pt]
 \item We propose a novel solution UNCOVER that enables unknown-awareness in object detection in real-time.
 It achieves strong recall of unknown objects on Cityscapes~\cite{cityscapes}, BDD100k~\cite{Yu2020BDD}, Fishyscapes~\cite{Blum2021Fishy}, and SegmentMeIfyouCan~\cite{Chan2021SMIYC}, with up to 25\% improved recall over YoloWorld and a very limited complexity increase.

 \item We further propose a depth-based post-hoc strategy to reduce false positive detections. This strategy is based on interpretable classical computer vision techniques and can be applied to any object detector. %
On average, we could reduce false the positive rate by 18.4\% while boosting recall by 4.1\%.

\item We extend anomaly segmentation benchmarks for evaluation. Specifically, we make segmentation masks usable for object detection and define a false positive metric based on region of interests, such as drivable area based on the road mask.
 
\end{itemize}

\section{Related Work}
\label{subsec:related}

\paragraph{Anomaly segmentation in AD}
To improve AD safety, one line of work is to perform anomaly segmentation, generating a binary mask for all unknown objects in the scene. Similar to semantic segmentation, the generated mask does not localize every unknown instance separately. As high-quality semantic segmentation typically relies on a heavy pixel decoder and high-resolution feature maps, anomaly segmentation building on top of standard semantic segmentation architectures, e.g., ~\cite{Tian2022pebal,Liu2023rpl,Rai2023Mask2Anomaly}, is typically less computationally efficient than object detection networks like YOLOs~\cite{bochkovskiy2020yolov4}. %
A more recent line of work on real-time panoptic segmentation~\cite{wu2022yolop,Zhan2024YoloPX} aim at including some segmentation like lane detection into the object detection while preserving the real-time performance. However, they still lack anomaly segmentation capability. %
Our method is based on object detection. Inspired by anomaly segmentation, we introduce an occupancy score to indicate if the predicted bounding box contains some parts of objects. As shown in \autoref{fig:occupancy}, the occupancy map can separate foreground things and background stuffs without the necessity of a pixel-wise segmentation and not being restricted to OOD detection on the road only.

\paragraph{OOD detection in object detection}
Object detection targets two tasks, i.e., classification and localization. Unknown objects may still be localized. Without unknown awareness, they will be mapped to the training classes, degrading the average precision of the known classes~\cite{Dhamija2020Elephant}. Therefore, one line of work focuses on avoiding such misclassification, leveraging image-level OOD detection techniques to better separate novel objects from the training classes in the classification head of object detectors~\cite{Du2022vos,Du2022siren}. We focus on a different challenge which is to improve the recall of unknown objects, as they may not even be localized in the first place. Misclassifying an unknown risky object as one of traffic participant classes may still result in similar planning and decision making. Overlooking them can be more critical. 

\paragraph{Open-world object detection}
In response to the closed-world assumption, the field has seen a pivot towards open-world object detection~\cite{joseph2021towards,Konan2022FCOSopen,Dhamija2020Elephant,Du2022vos,Liang2023sniffer,Zohar2023prob,Gupta2022owdetr,kim2021oln,Wang2023RandomBox,Huang2022Good}, which focuses on
incrementally learning new objects~\cite{joseph2021towards,Zohar2023prob,Gupta2022owdetr} in the given data.
However, unknown objects might be so different to the original data that generalizing inside the given data is not enough. %
Moreover, prior work often adopted two stage object detectors such as Faster-RCNN~\cite{Ren2015}, while some more recent work changed to transformer-based architectures such as DETR~\cite{carion2020end}. However, both architectures are not suitable for current real-time systems. The work~\cite{Konan2022FCOSopen} extended FCOS~\cite{Tian2019fcos} (a one-stage anchor-free object detector) for open-world object detection, using the localization quality scores initially proposed by~\cite{kim2021oln}. Similar to~\cite{Konan2022FCOSopen}, we aim at real-time solutions. We propose a novel objectness measure via occupancy prediction rather than using localization quality measures, as the latter is biased towards the known objects.

\section{Method}
\label{sec:method}

\picLarge{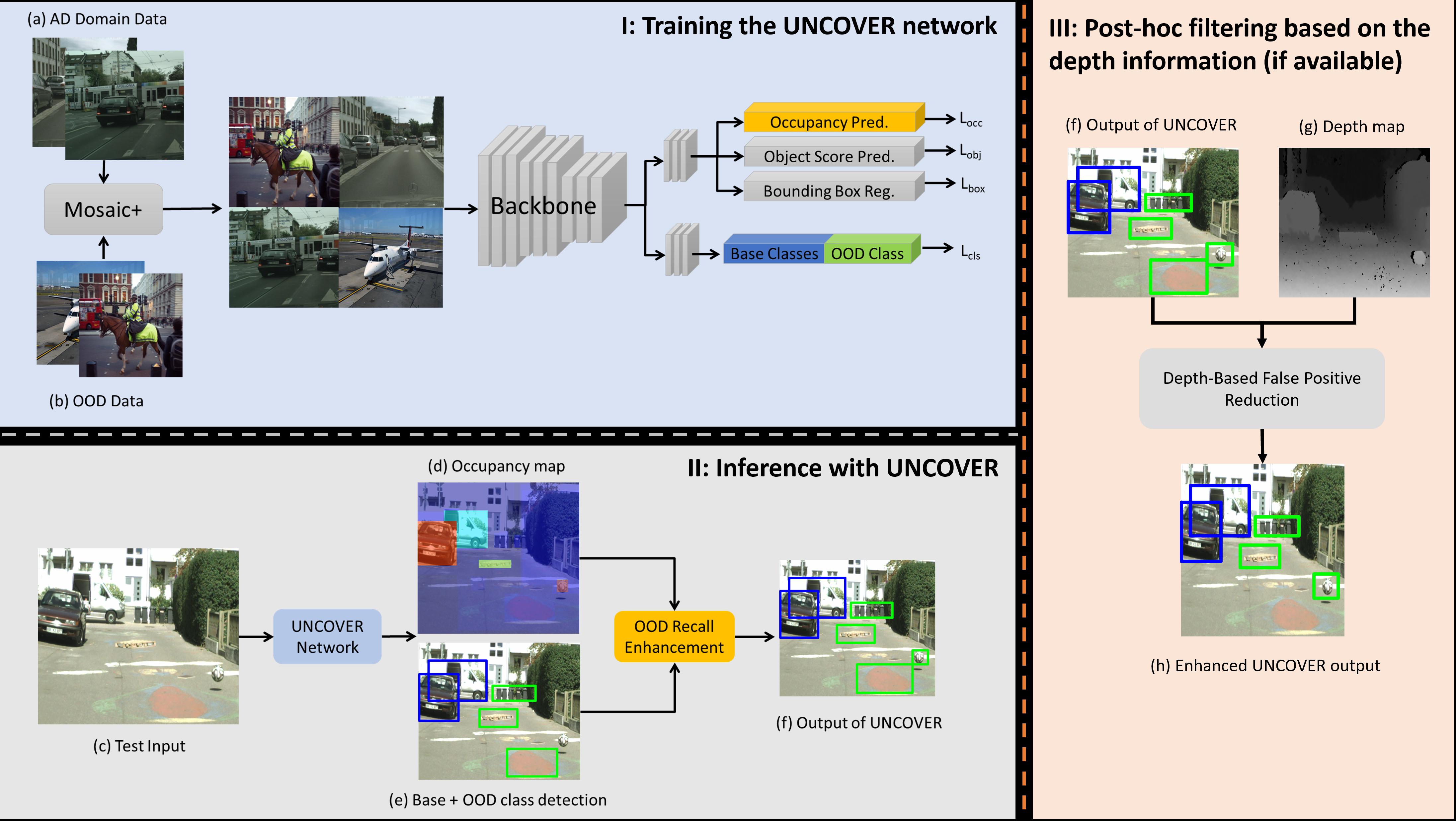}{UNCOVER -- We enable unknown object detection by adding OOD data (via Mosaic+), one extra class for OOD classification (dark blue), and one regression output (yellow) to predict the occupancy of each detection, i.e. phase I at training.
In phase II, UNCOVER exploits the occupancy prediction to improve the recall of unknown objects in addition to the OOD class detection via the classification head. If depth information, commonly found in AD systems, is available, UNCOVER uses an interpretable filtering step to reduce false positive detections (see phase III). Note, in addition to Mosaic+, we also use Mixup; it is left out of the diagram for simplicity. 
}{0.95}

Our method, \textbf{UNCOVER}, is designed to equip real-time object detection systems with awareness of unknown objects. %
\autoref{fig:pipeline_1} illustrates the three main aspects of UNCOVER. %
The model contains one additional class and an occupancy prediction head, and is trained with a mixture of AD data and other domain data for improved object diversity at training. %
At inference time (Phase II), the newly learned occupancy prediction serves as a measure of objectness, improving the recall of unknown objects. %
Given that AD systems often provide depth information for a scene, we also propose a simple and interpretable depth-based filtering for false positive detection reduction, i.e. Phase III.

\subsection{Preliminary}
\label{subsec:base model}

We showcase our design on top of modern, one-stage, anchor-free object detectors due to their real-time capability and competitive performance. Taking YOLOX~\cite{Ge2021yolox} and YOLOv8~\cite{Jocher_Ultralytics_YOLO_2023} as examples: each has a backbone and neck network followed by two decoupled heads, i.e., one for classification and one for regression. The classification loss $L_{cls}$ is based on binary cross-entropy and averaged over all classes. Besides the regression loss $L_{box}$ for learning the bounding box coordinates, YOLOX has an additional prediction output noted as \emph{Obj.} score in the regression branch. It classifies each detection based on if it has a matched ground-truth bounding box; its training loss $L_{obj}$ is binary cross entropy based.
We describe our modification on top of this base architecture.

\subsection{Unknown Object Detection (Phase I and II)}
\label{subsec:uDet}

To achieve high recall of OOD objects, UNCOVER 1) leverages data from other domains with an OOD class, 2) trains an occupancy prediction and 3) filters based on the occupancy score.

\subsubsection{Extra OOD Class and Mosaic+}
UNCOVER introduces an extra class, termed OOD class, so that detections can be classified as "Unknown". As the training set does not involve any OOD annotations, we include other datasets~\cite{mscoco,Gupta2019Lvis} that have annotated objects beyond the training classes of the AD dataset. Note, there are also objects having the same semantic classes as the AD dataset. They will still be trained with the corresponding known class in the classification head. %
To seamlessly incorporate new data for training, we extend Mosaic, a strong data augmentation scheme initially introduced in YOLOv4~\cite{bochkovskiy2020mosaic}.
Mosaic concatenates multiple images into one, greatly improving the training data diversity.
Our modification is to fetch images from two different sources instead of one, i.e., Mosaic+ as shown in \autoref{fig:pipeline_1}. 
Using data from different domains introduces a domain gap, as highlighted by \cite{Zhang_2023_ICCV}, which can affect the efficacy of using external datasets like MS COCO\cite{mscoco} for training anomaly segmentation in AD contexts.
To bridge this gap, we employ Mixup after Mosaic+, blending the composed image with an AD scene image, leveraging techniques also used in YOLOX~\cite{Ge2021yolox}.

\picThree{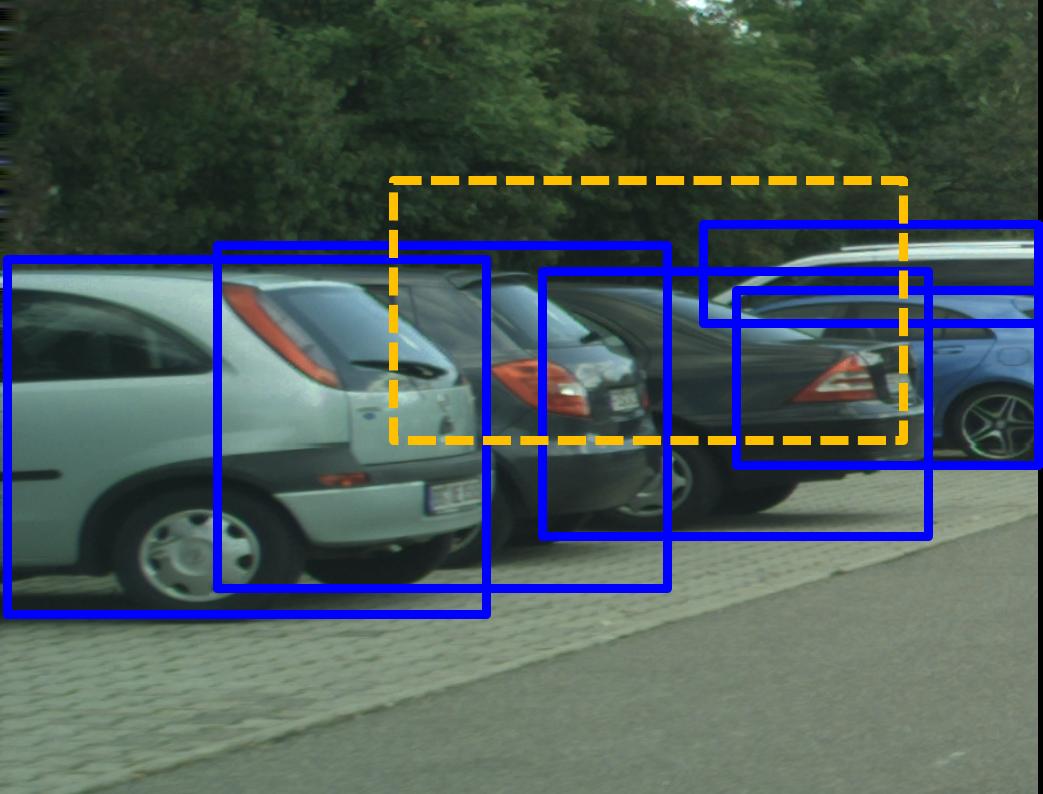}{$Obj = 0$}{$IoU = 0.4$}{$Occ = 0.85$}{
We compare three different objectness measures for ground truth (GT) bounding boxes (blue) and the predicted bounding box (dashed line, yellow).
The following scores are the target for optimization, not the actual output.
For (a) $Obj$~\cite{Ge2021yolox}, the score is one if positively matched to one GT box, or zero like in the presented case.
For (b) $IoU$~\cite{kim2021oln} the highest IoU achieved with one of the GT boxes. %
For (c) our proposed $Occ$ concerns the intersection with all GT boxes and thus does not rely on a valid matching, like a). Thus, even when the localization is difficult, occupancy with one or more objects is easily determinable. 
Moreover, $Obj$ and $IoU$ may require matching classes, while we are class-agnostic, allowing better generalizability to unknown objects.
}

\subsubsection{Occupancy Prediction}
\label{subsec:occupancy}

While MS COCO~\cite{mscoco} and LVIS~\cite{Gupta2019Lvis} have many more object classes than AD datasets, it is still unavoidable that the OOD class overfits to those specific classes in both datasets. Therefore, we introduce an occupancy prediction in the regression branch, which is supervised to evaluate the objectness in a class-agnostic manner. Then, we can decide whether to keep detections that have high scores even if their classification confidences may be low. 
For measuring objectness, IoU predictions are commonly used~\cite{kim2021oln,huang2023good}. In the context of YOLOX, that could be the \emph{Obj.} score~\cite{Ge2021yolox}. Both scores were developed for measuring the localization quality and may not respond well to unknown objects on which the model is expected to deliver lower localization quality. %

Instead of using a detection quality measure as the proxy for objectness, we introduce an occupancy prediction score. The score measures the ratio of the predicted area that overlaps with the ground truth bounding boxes. Specifically, the training loss $L_{occ}$ based on the binary cross entropy %
is formulated as:
\begin{equation}
\label{eq:loss}
\begin{split} 
    L_{occ}(b_{occ}) & = -t_ {occ} \cdot \ln(b_ {occ}) - (1-t_ {occ}) \cdot \ln(1-b_ {occ})\\
    t_ {occ} &= \frac{| b_{pred} \cap (\bigcup_{i=0}^{n} b_{{gt}_i} |) }{|b_{pred}|},
\end{split}
\end{equation}
where $t_ {occ}$ is the target and $b_ {occ}$ is the occupancy prediction. Here, $|\cdot|$ denotes the area in pixels, $b_{pred}$ the predicted bounding box, $\bigcup_{i=0}^{n} b_{{gt}_i}$ is the union of all ground-truth bounding boxes $(b_{{gt}_i})$ in the image. If that area is mostly covered by objects, the target $t_ {occ}$ gets closer to one. We also introduce an approximation for simplifying the computation of $t_ {occ}$ in the supp. material. 

As illustrated in \autoref{fig:class-agnostic}, the predicted bounding box (yellow dashed line) contains objects, so that the anchor point generating it shall contain object features, i.e., vehicles. The IoU and \emph{Obj.} score suppress their response to the object features in such a case. In contrast, the supervision signal for the occupancy prediction is stronger when the overlapping area is larger. A larger area indicates that the features yielding the bounding box prediction are mostly from objects.

\subsubsection{OOD Recall Enhancement}
After training, UNCOVER can generate detections with multi-class labels, including the OOD class. During inference, a standard filtering is applied to remove low-quality predictions which usually do not contain any objects of interest. In YOLOX, such filtering is based on the product of the maximum classification probability and the \emph{Obj.} score, i.e., $sco$. Only detections with $sco\geq\mu_{sco}$ are kept. As it is challenging for the model to achieve similar localization qualities for unknown objects as for known ones, we introduce the second filtering step for OOD recall enhancement. It retains the detections with ($sco<\mu_{sco}$) yet with a high occupancy score $occ\geq\mu_{occ}$. These retained ones will be considered as OOD objects in addition to those kept in the initial filtering step as OOD.

\subsection{Depth Based False Positive Reduction (Phase III)}
\label{subsec:dbf}

\picThree{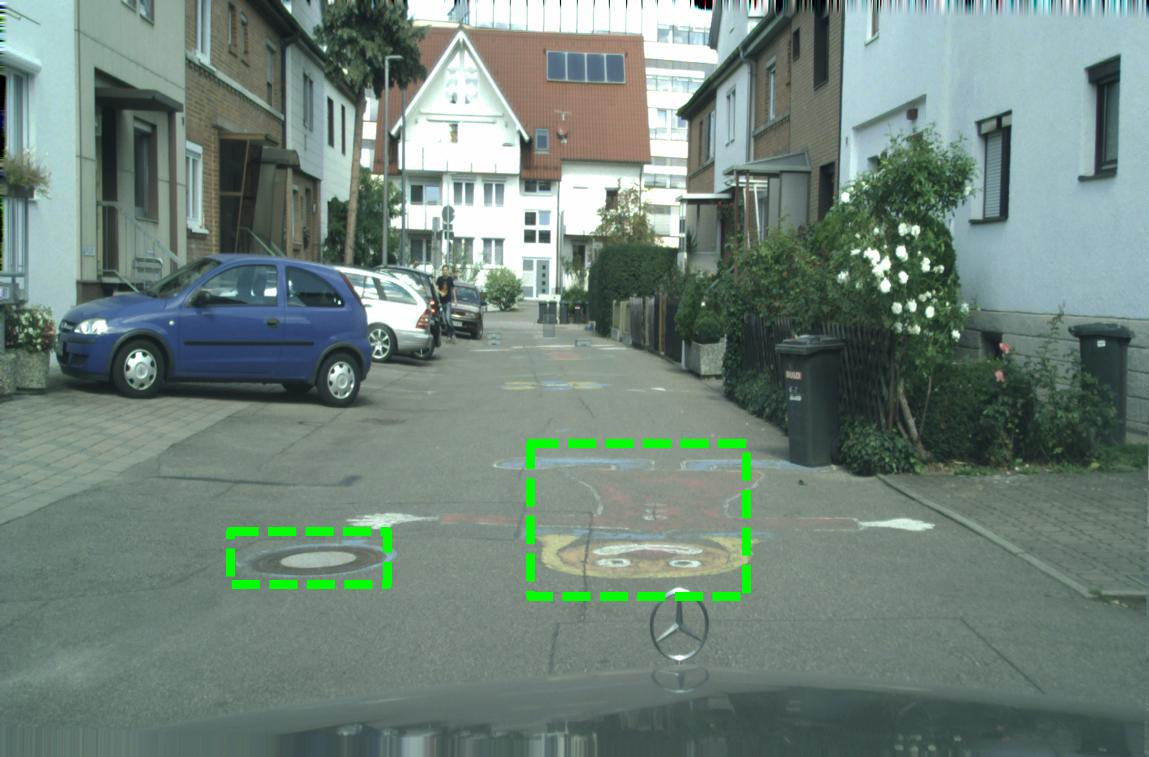}{Fishyscapes}{Cityscapes}{BDD100K}{Examples where visual cues are not robust, yielding false positive detections.}

Monocular inputs are cost-effective but can lead RGB-based detection to depend heavily on appearance cues, which may not be always reliable. \autoref{fig:false-positive} illustrates failure cases where non-objects are mistakenly identified due to their distinct textures from the background.
Such false positives, particularly on drivable surfaces near the vehicle, could  critically undermine the system reliability e.g. of emergency brakes.
To suppress such failures, we hypothesize the most effective solution is to leverage geometric cues, which provide a more holistic view and complement the appearance cues.
To verify our hypothesis and given the fact that depth information is often available in AD systems, we experiment on depth-based filtering. Note, here we only exploit the depth information for filtering the detections that are already generated by UNCOVER. %

The main idea is to measure the changes in depth within each bounding box. See for example the bird in \autoref{fig:dbse}, which is difficult to spot in the raw depth estimation but easier to discern in the change of depth. An indication of potential objects is having few local depth changes within the bounding box; on the other hand, the flat surface of the road exhibits continuous local depth changes. To detect such patterns, we devise Algorithm~\autoref{alg:dbse}, which is low complexity, interpretable, and purely post-hoc, without the need for retraining the network. Specifically, our algorithm involves pre-processing the depth map to highlight areas with depth changes by leveraging morphological transformations~\cite{fisher1996morph} and the Sobel operator~\cite{sobel1968}. The algorithm determines the likelihood of an object sticking out vertically from the ground by further evaluating the proportion ($c$) of pixels within a predicted bounding box that have depth change less than ten. Objects are characterized by having a few local changes. Therefore, the ones with $c\geq \mu$ are kept. %
While depth estimation quality becomes worse farther away from the camera, this does not affect us negatively. Since depth change can only be detected within a certain distance, the detections in far distance have $c$ close to its maximum and are always kept. The filtering mostly affects close objects. This is of practical interest, as closer objects are more relevant to the action for the very next step.

\picThree{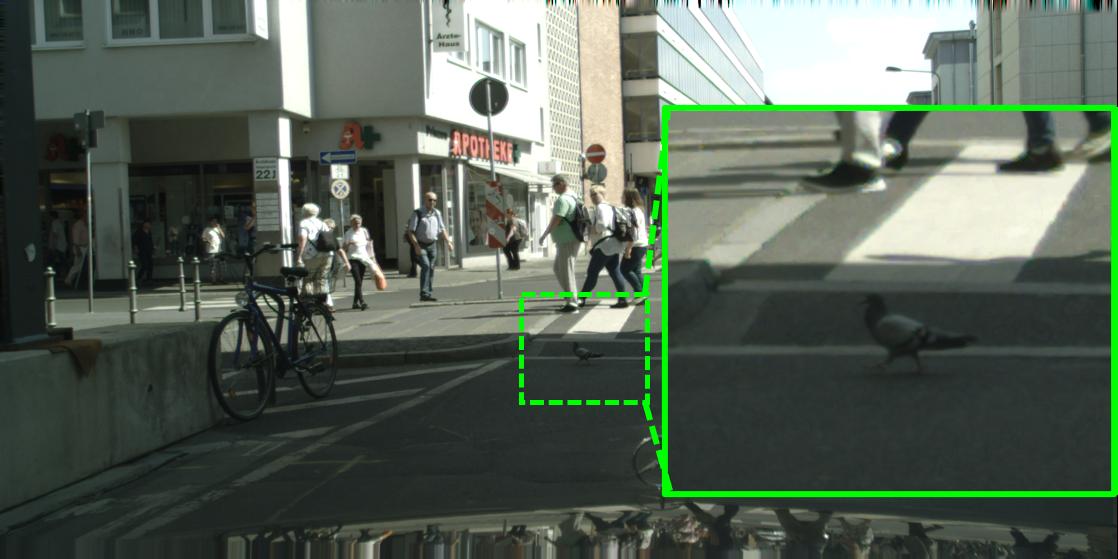}{RGB}{Depth}{Change of Depth}{Geometric cues from depth. For objects with geometric shapes such as the bird, it can be detected via the change of depth in the area. For road marking, there is no depth change. Therefore, depth can be exploited to filter non-objects in near range. 
}

\begin{algorithm}[tb]
\caption{Depth Based False Postitive Reduction}\label{alg:dbse}
\begin{algorithmic}
\Require Depth Image $D$, Predicted Bounding Box $x_1,x_2,y_1,y_2$
\State \text{\# Calculation of depth change $C$ from $D$}
\State $D \gets dilation(D,kernel=10)$ \Comment{Morphological Closing with dilation and erosion}
\State $D \gets erosion(D,kernel=10) $
\State $C \gets sobel(D, direction=y, kernel=5) $\Comment{Sobel operation in y-direction}
\\
\State  \text{\# Calculate acceptance / rejection of every predicted bounding box $x_1,x_2,y_1,y_2$}
\State $bbox \gets C[x_1:y_1,x_2:y_2]$ \Comment{Pixels of bounding box in $C$}
\State $c \gets count(bbox < 10)  / size(bbox)$ \Comment{Proportion with minimal depth change}
\If{$c \geq \mu$} 
\State Accept bounding box as potentially hazardous 
\Else
    \State Reject bounding box as non-hazardous
\EndIf
\end{algorithmic}
\end{algorithm}

\section{Experiments}
\label{sec:experiments}

\paragraph{Datasets}
\label{subsec:datasets}
We consider two standard AD object detection datasets for training, i.e., Cityscapes (CS)~\cite{cityscapes} and BDD100k~\cite{Yu2020BDD}. The former encompasses images of urban street scenes in Europe, while the latter was collected in U.S.. They both have eight annotated object classes for training: person, rider, car, truck, bicycle, train, bus, and motorcycle.
To facilitate learning to detect unknowns, we incorporate also images from COCO~\cite{mscoco} and LVIS~\cite{Gupta2019Lvis}. Their label space alignment with CS and BDD100k is detailed in the supplementary material.

For evaluation, we construct the unknown object annotations by improving the annotations of CS on the class 'dynamic', 'trailer', and 'caravan'.
The 'dynamic' class is described as 'Things that might not be there anymore the next day/hour/minute: Movable trash bin, buggy, bag, wheelchair, animal'\footnote{\url{https://www.cityscapes-dataset.com/dataset-overview/}}, with examples in the appendix.
We derived bounding box information from semantic masks using blob detection for the non-instance-specific 'dynamic' annotation, with details in the supplementary material.
BDD100k provides object detections for the classes 'traffic sign' and 'traffic light' which we do not use during training and thus can be used as unknowns during evaluation. %

Besides CS and BDD100k, we also benchmark on Fishyscapes Lost and Found (FS L\&F)~\cite{Blum2021Fishy} and Segment Me If You Can (Anomaly / Obstacle Track)\cite{Chan2021SMIYC}. %
They have real OOD scenarios, but without bounding box annotations.
Therefore, we process them to generate the annotations for benchmarking unknown object detection methods, and refer to the supplementary material for details. 

\paragraph{Evaluation Metrics}
We consider three metrics: mean Average Precision (mAP in \%)~\cite{mscoco}, Recall (R@100 in \%) and False Positive Rate (FPR@100 in ‰) at the IoU threshold $0.5$. mAP is measured on the training classes, i.e., knowns. R@100 and FPR@100 measure the success rate of unknown detection, where the top $100$ unknown detections are kept for evaluation. Except the objects from the training classes, unknown objects are not exhaustively annotated in the datasets; therefore we cannot directly measure mAP for them. FPR@100 is computed for a specific region of interest (RoI) where we have exhaustive annotations for every pixel. In our case, we use the "road" mask to derive the RoI and report FPR within that specific region. %
For run-time evaluation we measure the frames per second (FPS), see details in the supplementary material.

\paragraph{Implementation Details}
\label{subsec:details}

We adopted the YOLOX architecture with a CSPNetX backbone, configured to process images at a resolution of $640\times 640$ pixels, as our base objection detection model~\cite{Ge2021yolox}.
Our implementation is based on the mmdetection framework\footnote{\url{https://github.com/open-mmlab/mmdetection}} and their default parameters.
The experiments were mainly conducted on one  NVIDIA V100 with an average training time of 20 hours per experiment.%
The Mosaic+ augmentation strategy replaces two out of four images from an auxilary OOD dataset like MS COCO and LVIS~\cite{Gupta2019Lvis,mscoco}. %
Our occupancy loss is added to the original training loss of YOLOX with the weighting $w_o=1.0$.
To address the imbalance among Cityscapes/BDD100k classes, we applied class weights of [0.5,0.9,0.4,1,1,1,1,0.7,10], where the last entry for the extra OOD class is up-weighted due to the greater learning complexity. %
For thresholding, we used grid-searched optimized values of $\mu = 0.3$, and $\mu_{occ} = 0.01$ where $\mu_{sco} = 0.01$ is the default choice in YOLOX. For more details, see the supplementary material.

\subsection{Unknown Object Discovery via Occupancy Prediction}
\label{subsec:eval}

\tblResultsocc
To filter generated bounding boxes, state-of-the-art object detectors rely on objectness measures. Namely, a bounding box with a high objectness measure is likely to capture an object. In order to improve the recall of unknown objects, it is important to learn a generic sense of objectness, avoiding biases towards the training classes. \autoref{tbl:results_occ} compares three different objectness measures, where the base model is the same, i.e., YOLOX with Mosaic+. %
\emph{Obj.} score is a built-in prediction of YOLOX, and we further use it to do OOD recall enhancement. %
As \emph{Obj.} score also considers the class information (during SimOTA), it performs worse than the others, as the objectness measure should be class-agnostic, i.e., generalizable across knowns and unknowns. Compared to the IoU score which measures the localization quality, our occupancy measure is more responsive to unknowns, yielding higher recall across all benchmarks. %
Among the five benchmarks, Anomaly and Obstacle are less challenging as the objects are salient in the scenes. In contrast, the unknown objects in FS and BDD100k are rather small, e.g., lost cargo objects, and traffic signs. The lowest recalls are on the reserved unknown classes from Cityscapes. As shown in the supp. material, the objects are often small and occluded in the background without clear view, thus more challenging to detect. We refer to the supplementary material for more visual results on comparing the objectnesss scores. The effectiveness of OOD class and objectness score are also ablated there.

\subsection{Comparison with Anomaly Segmentation}
While segmentation and object detection are two perception tasks with different goals, we find it interesting to compare them for unknown detection. Firstly, anomaly segmentation in AD has received more attention than object detection. Secondly, both Fishyscapes~\cite{Blum2021Fishy} and SegmentMeIfYouCan~\cite{Chan2021SMIYC} are anomaly segmentation benchmarks. %
It is expected that a segmentation model can outperform our model but at the cost of real-time condition.
\autoref{tbl:Results_anoseg} compares UNCOVER with two competitive anomaly segmentation methods PEBAL~\cite{Tian2022pebal} and RPL~\cite{Liu2023rpl}. As they were trained on Cityscapes and use MS COCO as OOD data, we also switch from using LVIS to MS COCO for a fair comparison. The predicted mask of PEBAL and RPL are converted into bounding boxes with blob detection as elaborated in the supplementary material. Our recall performance is better than PEBAL while similar to RPL. The main differences are on Fishyscapes and Anomaly based on the visuals in supplementary. We hypothesize that the anomaly mask conversion yields smaller detections, which is helpful on FS with more smaller objects on the horizon, while hindering on Anomaly with large objects.
A clear advantage of UNCOVER is its much lower complexity, achieving ten times higher throughput than PEBAL and RPL while having similar or better performance.

We also compare with open-world object detection and open-vocabulary object detection methods in the supplementary material. Unlike anomaly segmentation, they were not proposed and tailored for AD domains. The open-world object detector PROB~\cite{Zohar2023prob} and OW-DETR~\cite{Gupta2022owdetr} do not already perform well on Cityscapes. %
Grounding DINO~\cite{Liu2023GDino,Zhao2024mmgdino} (GDINO) and YOLO-World~\cite{Cheng2024YoloWorld} are open-vocabulary methods that allow to detect a wide range of object classes with text prompts. We prompt large language models (LLMs) to generate potential object classes that may appear in driving scenes (but outside the AD training classes) and use them to detect potential unknown objects. GDINO performs often within 10\% of UNCOVER while not being real-time, while YOLO-World achieves real-time but has a lower Recall on all 5 datasets. We conclude that our solution UNCOVER excels at detecting unknown objects in real-time while preserving known class performance.

\tblResultsanoseg

\subsection{Depth-based False Positive Reduction (DFR)}
\label{subsec:depthCompare}
\tblDBSE

Our last experiments highlight the role of depth information in reducing false positives.
Despite UNCOVER's relative low FPRs (see supplementary), false positives can be further reduced using geometric cues, such as depth information.
For Cityscapes and Fishycapes stereo images are available.
While the focus of our work is on monocular object detection, \autoref{tbl:dbse} shows that  depth-map based filtering improves all prior methods, including UNCOVER (ours). 
Not only does the filtering reduce FPRs, but it also improves recall. Filtering out ghost detections prevents them from consuming the budget, i.e., the top $100$ detections, for recall evaluation.
We see that on average DFR reduces FPR@100 by 18.4\% and increases R@100 by 4.1\%.
Since stereo depth estimation is not always available we evaluate the impact of monocular depth estimation in the supplementary.

\section{Broader Impact \& Limitations}
\label{subsec:limit}

Autonomous driving has a major potential impact on our society. 
However, reliable and fast system especially in unexpected situations are key for save driving. 
Thus, we have focused on enabling existing real-time models with simple modifications. Our final performance is then naturally also limited by the base model. Our method can be applied with a new model, and a joint design may boost the performance further. Moreover, while our focus is not on data augmentation, exposing models to some OOD data during training has been an effective solution. However, not every possible real-world object has an equal chance to be traffic relevant, e.g., many objects from the "accessory" category in MS COCO used in this study. Thus, prioritizing objects that are more likely in AD scenes but hard to collect in real world is an interesting direction.
We see no direct harmful application of our proposed method for autonomous driving.

\section{Conclusion}

In this work, we present a novel method UNCOVER (unknown class object detection for autonomous vehicles in real-time). %
By integrating a novel occupancy prediction head and augmenting training on the OOD class with diverse datasets, our method achieves state-of-the-art performance on five different datasets under the contrast of real-time prediction, with up to 25\% improved recall. %
Additionally, our depth-based filtering technique, utilizing geometric cues, significantly reduces false positives by 18.4\% and improves recall by 4.1\%. This shows the effectiveness of our methods on addressing open-world AD challenges in real-time.

\bibliographystyle{splncs04}

\bibliography{lib}

\newpage

\appendix

\section*{Appendix / supplemental material}

The supplementary material of the main paper is structured as follows:

\begin{itemize}
    \item In Appendix \ref{supp:details}, more details about our method UNCOVER. 
    \begin{itemize}
        \item \ref{supp:loss} a low-complexity approximation of our occupancy prediction loss function.
       \item \ref{supp:filterOne} hyper-parameter selection for OOD recall enhancement.
        \item \ref{supp:filterTwo} hyper-parameter selection for depth based false positive reduction.
    \end{itemize}

    \item In Appendix \ref{supp:evaluation},  more experimental details.

    \begin{itemize}
      
        \item \ref{supp:datasets} Dataset setting and processing for benchmarking unknown object detection.
    \item \ref{supp:runtime} Runtime measurement and comparison. 
        \item \ref{supp:anomaly} Comparison with anomaly segmentation models.
        \item \ref{supp:openvoc} Comparison with open-vocabulary models.
    \end{itemize}

    \item In Appendix \ref{supp:results}, additional experiment results

    \begin{itemize}
    \item \ref{supp:openvoccomp} Comparison results with two open-vocabulary object detection models.
        \item \ref{supp:ow} Comparison results with two transformer-based open-world object detection models.
        \item \ref{supp:ablation} Ablation of the impact of the different parts of UNCOVER.
        \item \ref{supp:mono} Comparison results with monocular depth estimation for DFR.
        \item \ref{supp:qual} Object detection visualizations generated by our model (UNCOVER), YoloWorld~\cite{Cheng2024YoloWorld}, GDINO~\cite{Zhao2024mmgdino} and RPL~\cite{Liu2023rpl}. Additionally visual results of YoloPX~\cite{Zhan2024YoloPX}.
        \item \ref{supp:occ} Visuals of occupancy vs. object predictions across datasets.
        \item \ref{supp:reproduce} Full reproducibility results for \autoref{fig:dbse_table}.
    \end{itemize}
    
\end{itemize}

\section{More implementation Details of UNCOVER}
\label{supp:details}

\subsection{Approximation of the Loss Function for Occupancy Prediction}\label{supp:loss}
To speed up the loss computation and gradient computation, we resort to an approximation of the loss in \autoref{eq:loss}, used for training the proposed occupancy prediction. Specifically, the intersection between the predicted bounding box and the union of all ground-truth boxes is approximated by its upper bound

\begin{equation}
\begin{split} 
        | b_{pred} \cap (\bigcup_{i=0}^{n} b_{{gt}_i}) |  &=  | \bigcup_{i=0}^{n} (b_{pred} \cap b_{{gt}_i} )| \nonumber \\
   & \leq  \sum_{i=0}^{n}  | (b_{pred} \cap b_{{gt}_i} )|. \\
\end{split}
\end{equation}

As an upper bound, the approximation potentially overestimates the intersection. It becomes tight when there are no or very few occlusions among the ground-truth bounding boxes. Empirically, we verified that this approximation works as effectively as the original loss formulation, while also reducing the training time. 

\subsection{Ablation on the filtering mechanism for OOD recall enhancement}
\label{supp:filterOne}
One of our core contributions is the introduction of the occupancy score and using it to improve the recall of OOD objects.
While the training loss aims at learning generic objectness, the learned occupancy score can be used in  multiple ways.
Aiming at real-time processing, we favor simple threshold-based solutions, namely, filtering detections based on real value comparison to the occupancy score. 
Through our ablation study, we find that the final performance is not sensitive to the exact threshold value $\mu_{occ}$ when it is in a reasonable region.
We use the same value as that used by YOLOX \cite{Ge2021yolox} for filtering objects from the training classes, i.e., $\mu_{occ}=\mu_{sco}=0.01$. It is kept the same for all tested benchmarks.
In \autoref{supp:occ} we show multiple visual examples that highlight how occupancy is more salient for objects than the $obj$ metric of \cite{Ge2021yolox}.
Especially for highly occluded or difficult instances, our occupancy prediction can more robustly identify the existence of such objects.

\subsection{Ablation on the filtering mechanism for depth-based FPR reduction}
\label{supp:filterTwo}
To further reduce the false positive rate, we use filtering with a single threshold setting $\mu$. As shown in \autoref{fig:dbse_table} (for numbers see \autoref{tbl:DepthCS} and \autoref{tbl:DepthBDD}), the performance is not sensitive to $\mu$. Our selection is based on a grid search on a hold-out validation set. 
In general, one can analyze the depth changes of the annotated training class objects and derive the threshold, as such geometry cues are rather class-agnostic. 
Moreover, we only need to verify whether the bounding box contains any object; thus this process is not sensitive to slight depth changes.

\section{More Experiment Details}
\label{supp:evaluation}

\subsection{Datasets and evaluation}
\label{supp:datasets}
As we use MS COCO~\cite{mscoco} and LVIS~\cite{Gupta2019Lvis} as auxilary data to augment the training on the AD datasets, Cityscapes~\cite{cityscapes} and BDD100k~\cite{Yu2020BDD}, their label spaces must be aligned; some classes from COCO and LVIS share the same semantic concepts as the training classes of Cityscapes and BDD100k, including for instance cars and persons.
Other classes do not share the same semantic concepts of the 8 training classes and we merge them into an 'OOD' class.

We further note that while 'traffic sign' and 'traffic light' are common training classes for AD, Cityscapes only provides semantic instead of instance masks for these.  Thus we removed them from the known classes during training, and used them for OOD recall and false positive evaluation.
Furthermore, we use the 'dynamics' class in Cityscapes for OOD evaluation. 
Two examples are given in \autoref{fig:cs-unknowns}.
For the classes for which only semantic masks are available, we extracted bounding boxes from the semantic masks with blob detection using the Python version of OpenCV's 'findContours'. In principle this could have led to either merging or splitting of instances depending on the connectivity of the mask.
We found this not to be an issue in practice since most samples are individual objects which do not overlap, or are separated by other classes.

For FPR@100 evaluation, we focus on the road area, as this is most relevant to driving safety.
Moreover, it is important that in the region of interest we need pixel-wise annotations to ensure that there are no un-annotated objects. %
For the object detection benchmark of BDD100k, the ground-truth road mask annotation is only available for a subset of the training set. %
Therefore, for our Cityscapes trained model, we use the BDD100k training set to evaluate FPR. When  training on BDD100k, we do not report FPR numbers due to the absence of road masks in the validation set. For Segement Me If You Can, we use the provided in-distribution masks to determine the ROI for FPR calculation. For the obstacle track this corresponds to the street surface as well. %
For the Anomaly Track, everything which is in-distribution for the original semantic segmentation mask is marked as a region of interest.

\picTwo{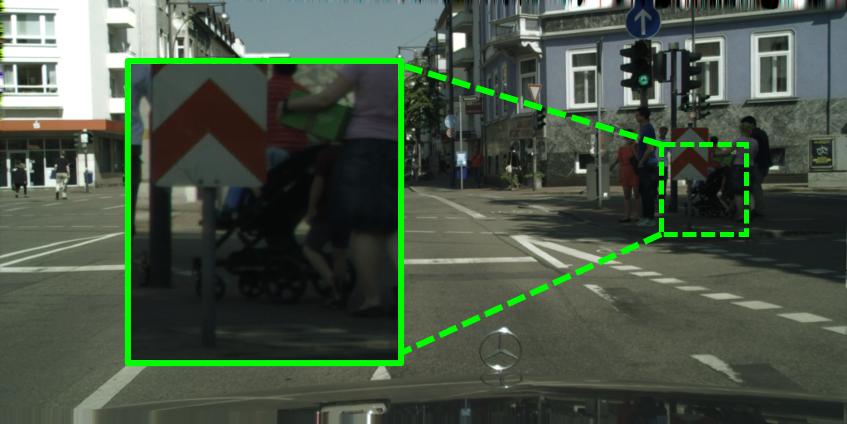}{Buggy}{Road signs}{Two examples of unknowns in Cityscapes from the 'dynamic' class}

For obtaining the depth information, we tried both stereo inputs and monocular depth estimator, i.e., a Global-Local Path Network~\cite{Kim2022GPN} specifically fine-tuned on the NYUv2 dataset~\cite{Silberman2012NYU}.
We mainly report results based on stereo depth estimation~\cite{konolige2008stereo} but show in \autoref{supp:mono} also results for the described off-the-shelf monocular depth estimator.%

\subsection{Runtime evaluation}
\label{supp:runtime}

We focus in our evaluation on comparing magnitudes of differences in runtime.
Overall, our runtime evaluation aims at ranking each model from the perspective of run-time capability; the exact computation cost and hardware complexity are beyond the scope of the evaluation. %
The run time evaluation was conducted on a NVIDIA V100 and a NVIDIA RTX2070. %
We calibrated the performance difference between the used hardware and  %
the reported FPS is based on averaged run-times across 100 repetitions. %
Since we evaluated based on reported system times, it can be expected that the numbers slightly change based on the remaining load of the system. This is not an issue for the reported results since the major differences are a magnitude slower or faster.

\subsection{Conversion of anomaly semantic masks into bounding boxes}
\label{supp:anomaly}
The evaluation metrics commonly used by the anomaly segmentation methods in the literature are threshold free. However, for practical usage, the per-pixel OOD scores generated by the anomaly segmentation methods need to be mapped into a binary mask, classifying each pixel as either OOD or not OOD. As the methods in the literature do not provide a way of setting such thresholds, we devise the following strategy to generate (instance-wise) OOD detections based on the per-pixel OOD scores. We try multiple thresholds and fuse their results. Specifically, for each threshold we determine which parts of the anomaly scores are above the threshold and use blob detection on the thresholded image.
The score for each detected blob / bounding box is the used threshold, since it describes the lowest confidence in the anomaly scores for a connected patch.
Thus, when using lower thresholds more parts of the image are considered to be unknown and larger bounding boxes are predicted.
Visual examples for converted RPL~\cite{Liu2023rpl} detections are shown in \autoref{supp:qual}.

\subsection{Open Vocabulary Prompting}
\label{supp:openvoc}

When using open-vocabulary models (such as GDINO and YOLO-World) for OOD-aware object detection, we generate text-based prompts in the following manner. For the known classes, we directly use the class names, i.e., 'person', 'rider', 'car', 'truck', 'bus', 'train', 'motorcycle', 'bicycle'. For unknown classes, it is impossible to be exhaustive. Nevertheless, it is possible to query LLMs such as ChatGPT~\cite{OpenAI2022ChatGPT} to generate the most AD-relevant classes beyond the training classes. Moreover, we make use of the definition of 'dynamic' class provided in Cityscapes, as they are actual objects which have already been observed in AD scenes. The final prompting texts for unknown objects include 'traffic sign', 'traffic light', 'buggy', 'road obstacle', 'construction sign', 'wheel chair', 'animal', 'trash bin', 'wheel chair',  'pallet', 'wheel', 'baggage', 'traffic cone', 'box', 'branch', 'bicycle', 'ball', 'toy', 'dog', 'bird', 'skateboard', 'scooter','cat', 'stroller', 'bench', 'fence', 'puddle', 'pothole', 'manhole cover',  'flower pot', 'bollard', 'roadwork barrier', 'fallen sign', 'shopping cart', 'ladder', 'sandbag', 'construction equipment', 'temporary road sign', 'street art installation', 'lost clothing', 'spilled cargo', 'advertising board', 'fire hydrant', 'electric scooter', 'table', 'chair', 'sun shade', 'helicopter', 'airplane', which are used in addition to the 8 training class names in above. 

Note, 'traffic sign' and 'traffic light' are treated as unknowns for the reason we provided in \autoref{supp:datasets}. Moreover, the class names are not mutually exclusive, e.g., animal and dog are both in the name list. This is because the open-vocabulary model reacts to each text prompt differently. Both animal and dog can become a traffic participant, creating hazardous road situations. In our setup, predicting either of them leads to an identical predicted label, i.e., 'OOD' object. 

\section{More results}
\label{supp:results}

\tblResultsopenvoc

\subsection{Comparison with open-vocabulary object detection}
\label{supp:openvoccomp}

VLMs enable detecting unseen unknowns by specifying names of potential traffic-relevant objects beyond training classes, allowing open-vocabulary detectors to identify them at inference. %
Grounding DINO (GDINO)~\cite{Liu2023GDino,Zhao2024mmgdino} demonstrated a strong generalization across different benchmarks. YOLO-World~\cite{Cheng2024YoloWorld} is a very recent work with the real-time capability. 
By querying GPT~\cite{OpenAI2022ChatGPT}, we generated 57 highly traffic relevant object classes beyond the training classes, e.g., buggy, trash bin or animal (details in supp. material) for GDINO and YOLO-World.
In~\autoref{tbl:Results_openvoc}, we observe that ours outperforms YOLO-World with regard to all recalls, e.g., up to 25\% improved recall on FS. The real-time capability is on a par. The more complex and thus slower GDINO surpasses our method in most cases but we are often at least within 10\% of their result showing similar performance.

We have also shown UNCOVER trained on both Cityscapes and BDD100k, which have a domain gap. The generalization from BDD100k to Cityscapes is much better than the opposite direction due to the larger size of BDD100k. Due to exposure to very diverse data at pre-training, the open-vocabulary models also generalize better than UNCOVER (CS) on BDD100k.

\subsection{Comparison with Transformer-based Open-world Object Detectors}
\label{supp:ow}

The recent trend in open-world object detection moves towards using transformer-based architectures. These models are typically trained with different losses and configurations than the CNN based ones. While transformers have great potential, and display advantages over CNN-based ones, they are currently still comparatively less data efficient at training and more expensive at inference time. 

In \autoref{tbl:ow}, we report additional results with models such as PROB~\cite{Zohar2023prob} and OW-DETR~\cite{Gupta2022owdetr}, which are much slower than YOLOX and YOLO-World. On the performance side, they are also inferior. We hypothesize that their transformer-based architectures require much more training data than Cityscapes. So far, their strong open-world detection performance was demonstrated on object detection benchmarks such as MS COCO, which has a lot more training samples than Cityscapes.

\tblow

\subsection{Ablation study on occupancy prediction vs. OOD classification}
\label{supp:ablation}

\tblAblations

We introduce two architecture modifications for UNCOVER, i.e., extra OOD class and occupancy prediction. From \autoref{tbl:Ablations}, we can observe that removing the classification head leads to a severe drop in the recall.
The reduction of the FPR is a symptom of less detected objects overall.

Removing occupancy prediction for OOD recall enhancement compromises both FPR and recall. These observations indicate that both components contribute to the success of UNCOVER. It is also worth noting that UNCOVER preserves the mAPs on the known classes, compared to the base model YOLOX. 
Based on \autoref{tbl:Ablations} and \autoref{tbl:Results_anoseg}, it is more beneficial to use LVIS over MS COCO as the data source for OOD-awareness training. LVIS has a much larger object diversity, which is a key factor for learning generic sense of objectness. 
More details and results can be found in the supplementary material.

\subsection{Depth-based False Positive Redcution with Monocular Depth Estimation}
 \label{supp:mono}

As stereo inputs are not always available, we further study the influence of depth estimation quality on the performance. For comparison, we take an off-the-shelf depth estimator Global-Local Path Network~\cite{Kim2022GPN} to generate a depth map, expected to be of lower quality than the one obtained from the stereo inputs. From \autoref{fig:dbse_table}, we can observe the benefit of high-quality depth information includes less sensitivity to the threshold setting $\mu$ in Algorithm~\autoref{alg:dbse}, and more pronounced gains in FPR and Recall.
A lower quality depth map leads to a more erratic and premature performance decline as thresholds.

\picTwoLegend{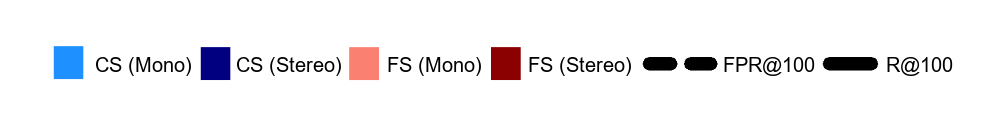}{Trained on CS}{Trained on BDD}{\emph{Relative} changes in FPR@100 and R@100 (Y-Axis) when applying different thresholds ($\mu$ at X-Axis) for depth-based filtering on top of UNCOVER -- Lower depth estimation accuracy (the monocular depth estimator~\cite{Kim2022GPN} in this case vs. stereo depth estimation) leads to increased sensitivity in determining the threshold $\mu$.
}

\FloatBarrier

\subsection{Qualitative Evaluation}
\label{supp:qual}

We report more visual results in \autoref{fig:fs_1}, \autoref{fig:fs_2}, \autoref{fig:anomaly_1}, \autoref{fig:anomaly_2}, \autoref{fig:obstacle_0} and \autoref{fig:obstacle_1} for our method UNCOVER, YOLOWorld~\cite{Cheng2024YoloWorld}, GDINO~\cite{Zhao2024mmgdino} and RPL~\cite{Liu2023rpl}.
The results are best viewed digitally.
We see that our method highlights unknown objects with high confidence while preserving predictions of known classes.

As a comparison we also provide in \autoref{fig:pano} a visual comparison to Panoptic Segmentation~\cite{Zhan2024YoloPX} under the real-time constraint.
Since no OOD or anomaly detection is integrated into the segmentation task, OOD objects are randomly detected or not on the driveable surface. 
The images also highlight why a direct comparison is not possible since objects like the giraffe are only partially on the driveable area and thus not fully segmented.

\picSixPreds{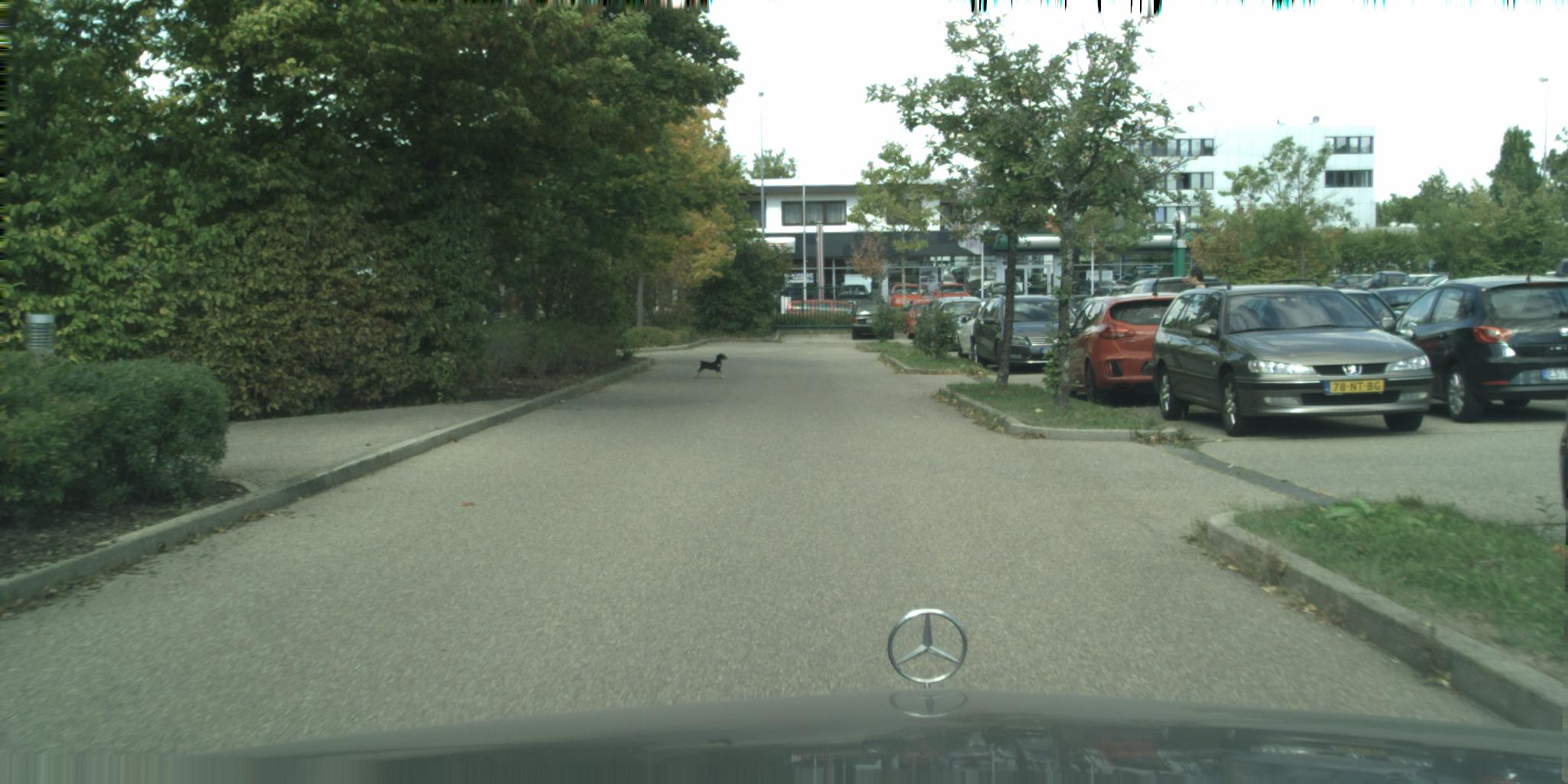}{Comparison of different methods for OOD-aware object detection on Fishyscapes. UNCOVER detects unknown objects, while maintaining high known class prediction accuracy. At the chosen threshold YOLOWorld correctly detects the OOD object, but misses known classes. GDINO and RPL (used here for object detection) suffer from many false positives.}
\picSixPreds{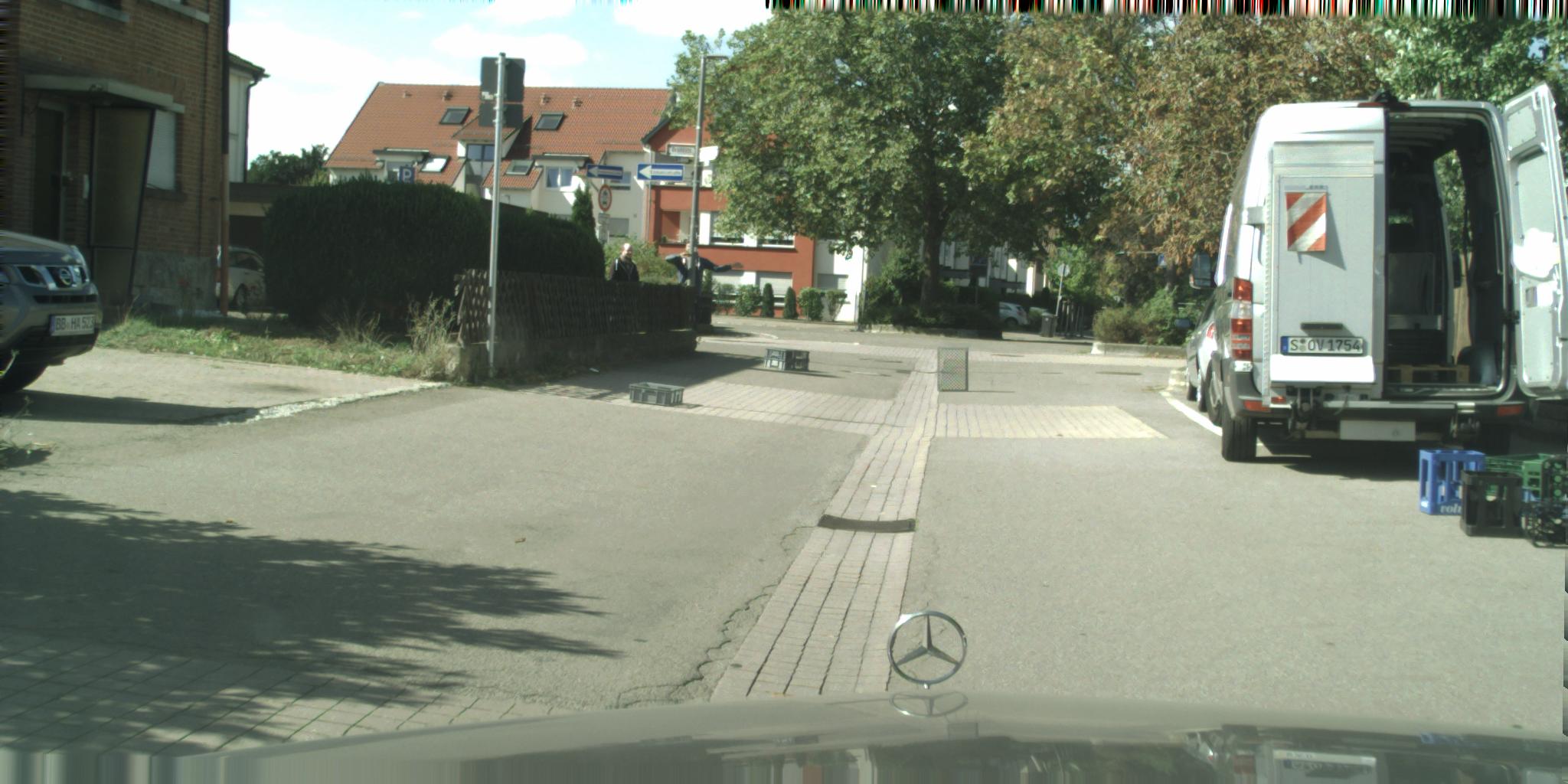}{Similar observations to \autoref{fig:fs_1}. It is also interesting to note that Fishyscapes only annotates unknown objects on the road, i.e., lost cargos. However, many detections made by UNCOVER in the background are actually also objects, even if they are neither part of the training classes nor of the labeled OOD objects in Fishyscapes. Therefore, we evaluate average recalls on unknown object detection. Average precision would penalize the right object detections due to lack of exhaustive object annotations.}
\picSixPreds{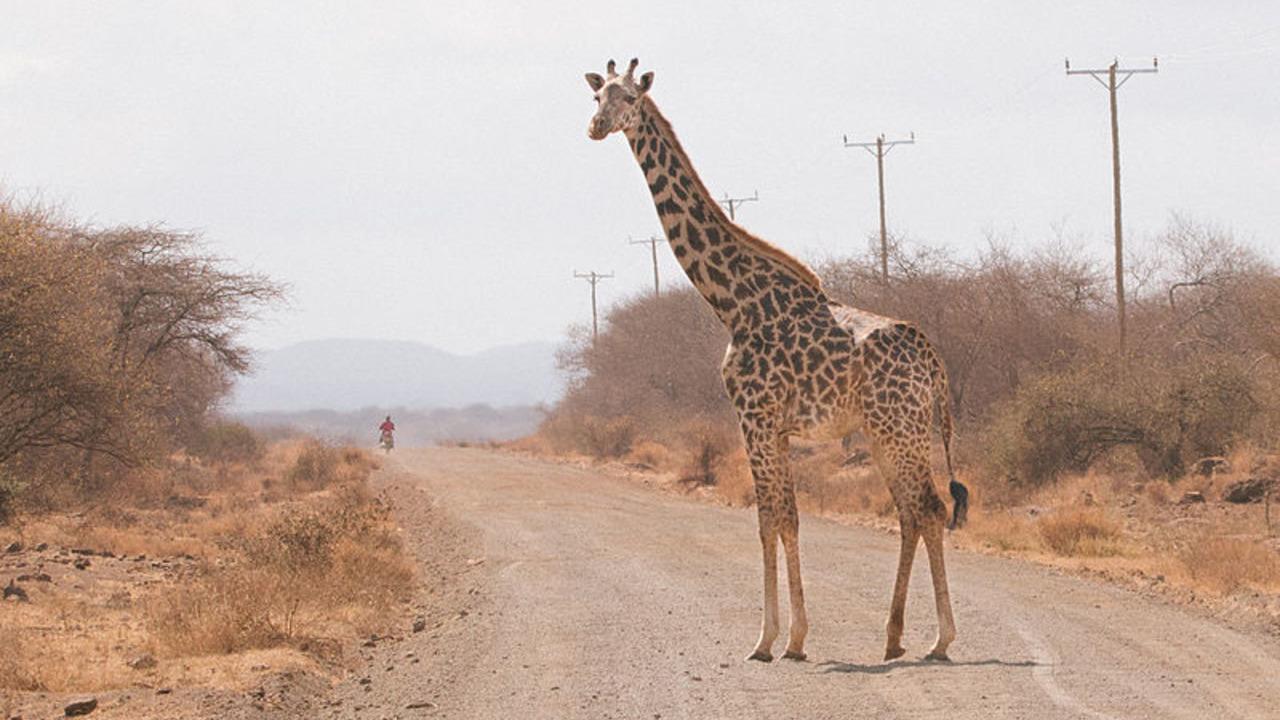}{Anomaly: In this case, the giraffe is a salient object which is rather easily  detected by all models. RPL has many small detections with poor localization quality. In contrast, UNCOVER detects background objects well in addition to the giraffe, even though they are not annotated.}
\picSixPreds{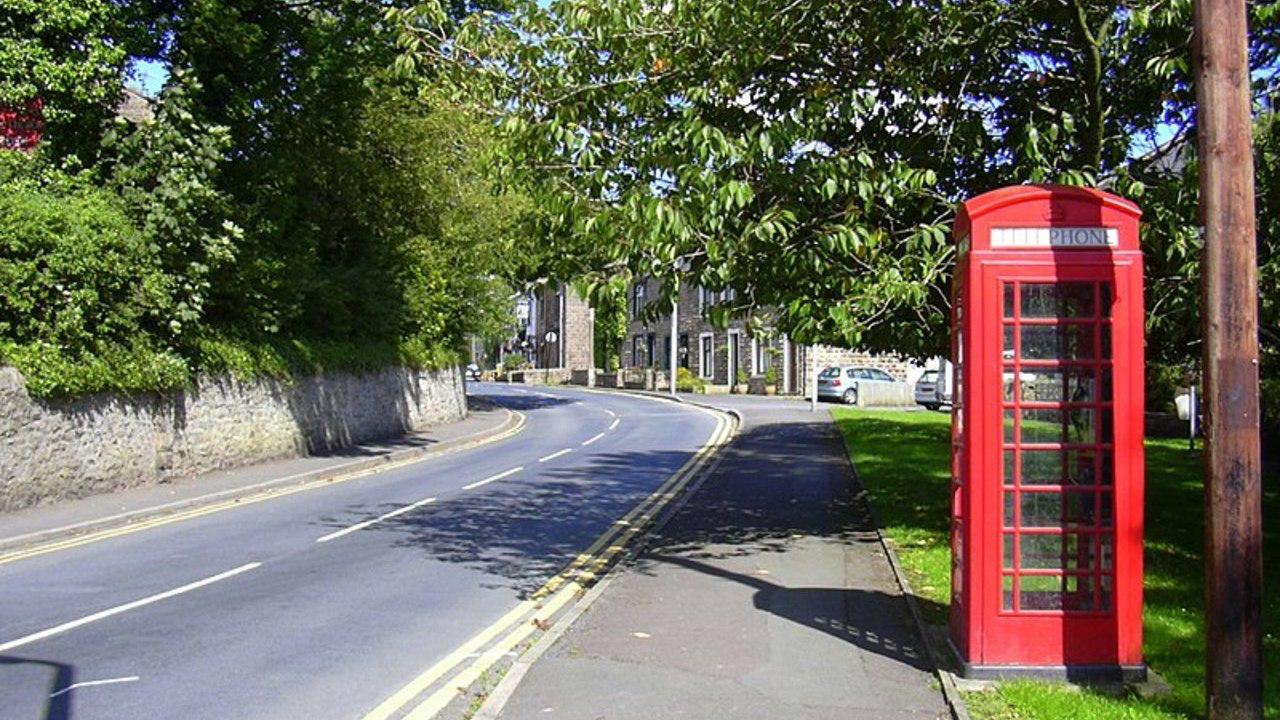}{Anomaly: In this example, both GDINO and UNCOVER have a ghost detection, i.e., tree shadow, whereas YOLOWorld has less positive detections on the background objects. RPL still suffers from many small detections. To remove the ghost detection, our proposed depth-based filtering can be applied.}
\picSixPreds{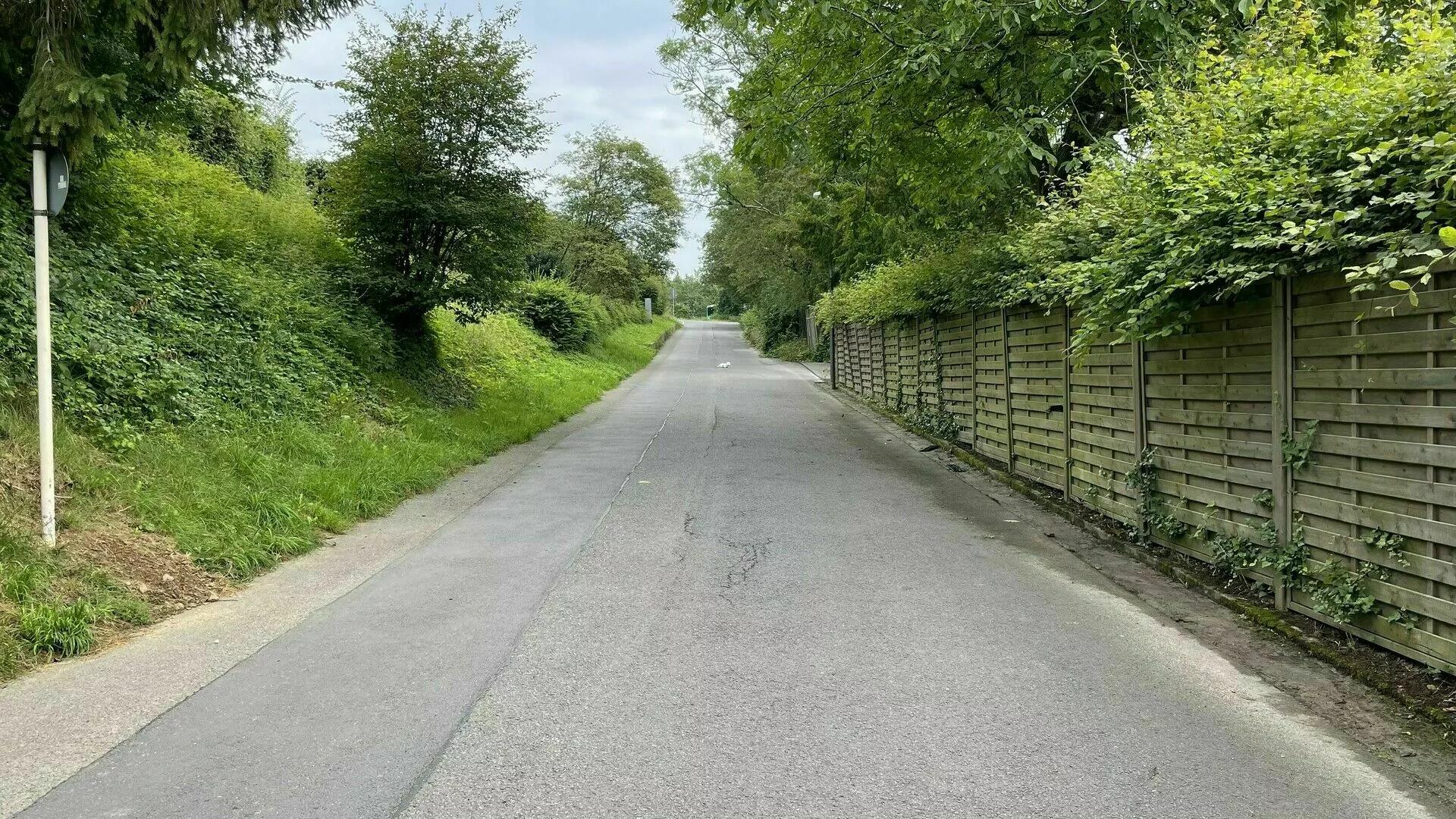}{Obstacle: this example demonstrates that UNCOVER is also capable at detecting further away small objects, that are either missed by YOLOWorld or detected at a much larger false positive rate, as shown for RPL.}
\picSixPreds{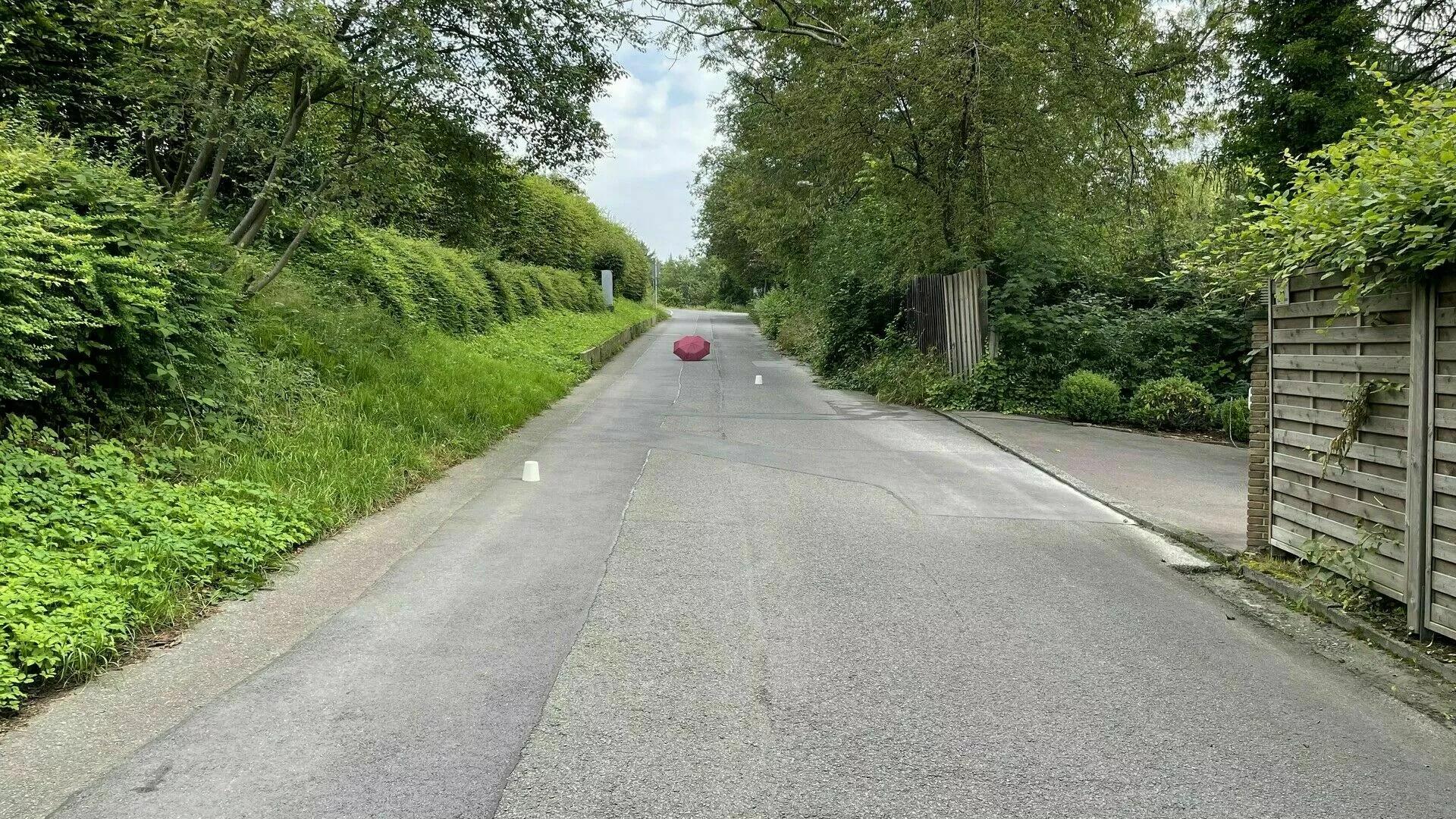}{Obstacle: Similar to the previous example. However, YOLOWorld detects the largest object while not picking up the others. }

\picSix{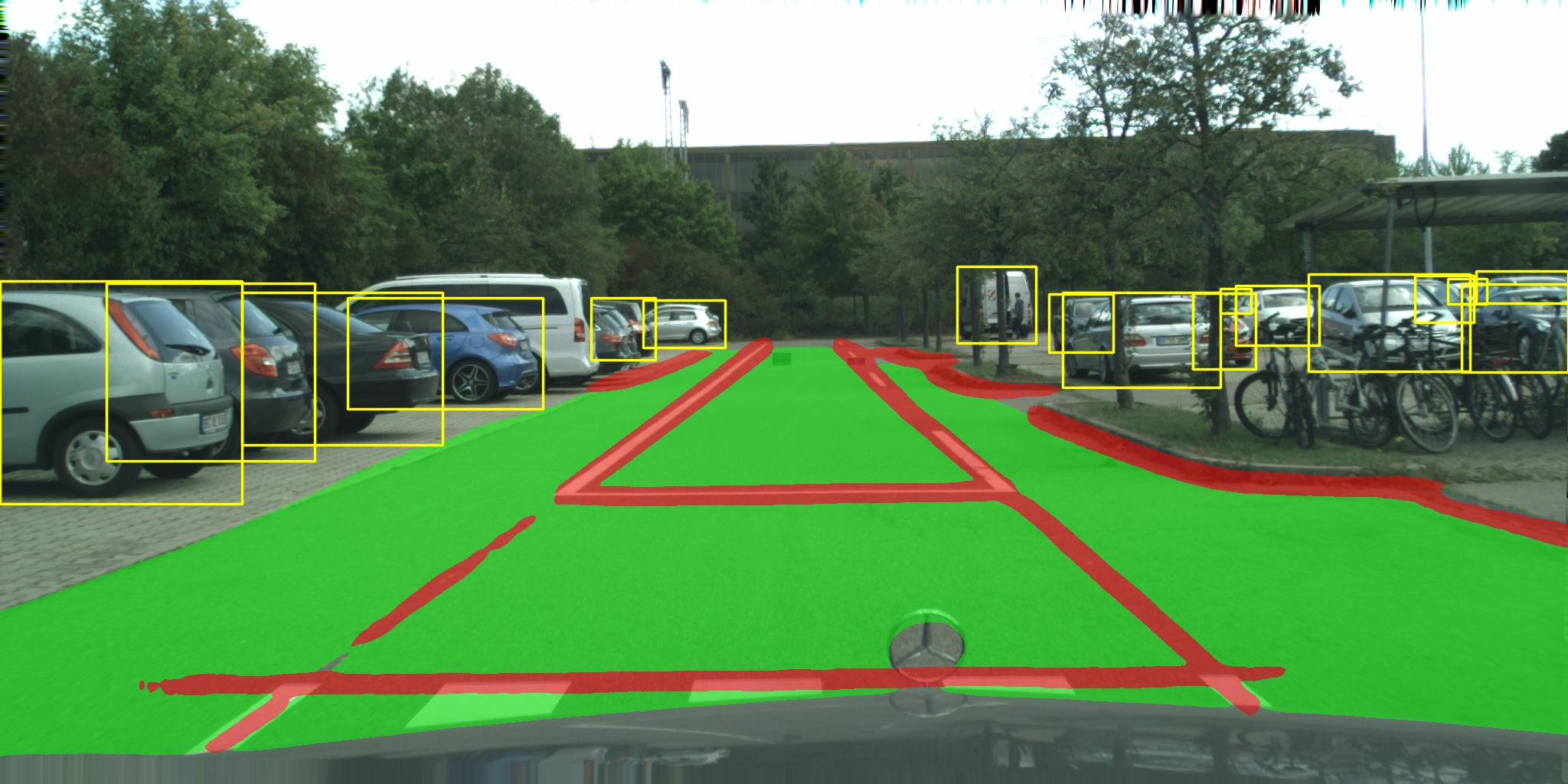}{Fishyscapes}{Obstacle}{Anomaly}{Visual results of YoloPX~\cite{Zhan2024YoloPX} across three different anomaly segementation tasks -- We see that the drivable area is sometimes aware of OOD objects like the ball and the giraffe. However it is not even consistent for the ball on the obstacle dataset.  }

\subsection{Occupancy visualization}
\label{supp:occ}
We also include three comparisons between $obj$ and our $occ$ prediction for all datasets in \autoref{fig:obj-cs},\autoref{fig:obj-fs}, \autoref{fig:obj-bdd}, \autoref{fig:obj-anno} and \autoref{fig:obj-obst}.
We see that $occ$ provides a more reliable indication of the objects, including both unknowns (e.g., ball in \autoref{fig:obj-fs-3})) and knowns (e.g. group of persons in \autoref{fig:obj-cs-3}). In contrast, the \emph{obj} score as a measure for localization quality has trouble indicating objects when the localization is not of high quality, e.g., a group of persons due to high occlusion or unknown objects due to lack of direct supervision. However, for those objects, it is safer to detect them as objects rather than ignore them as background.

\picSix{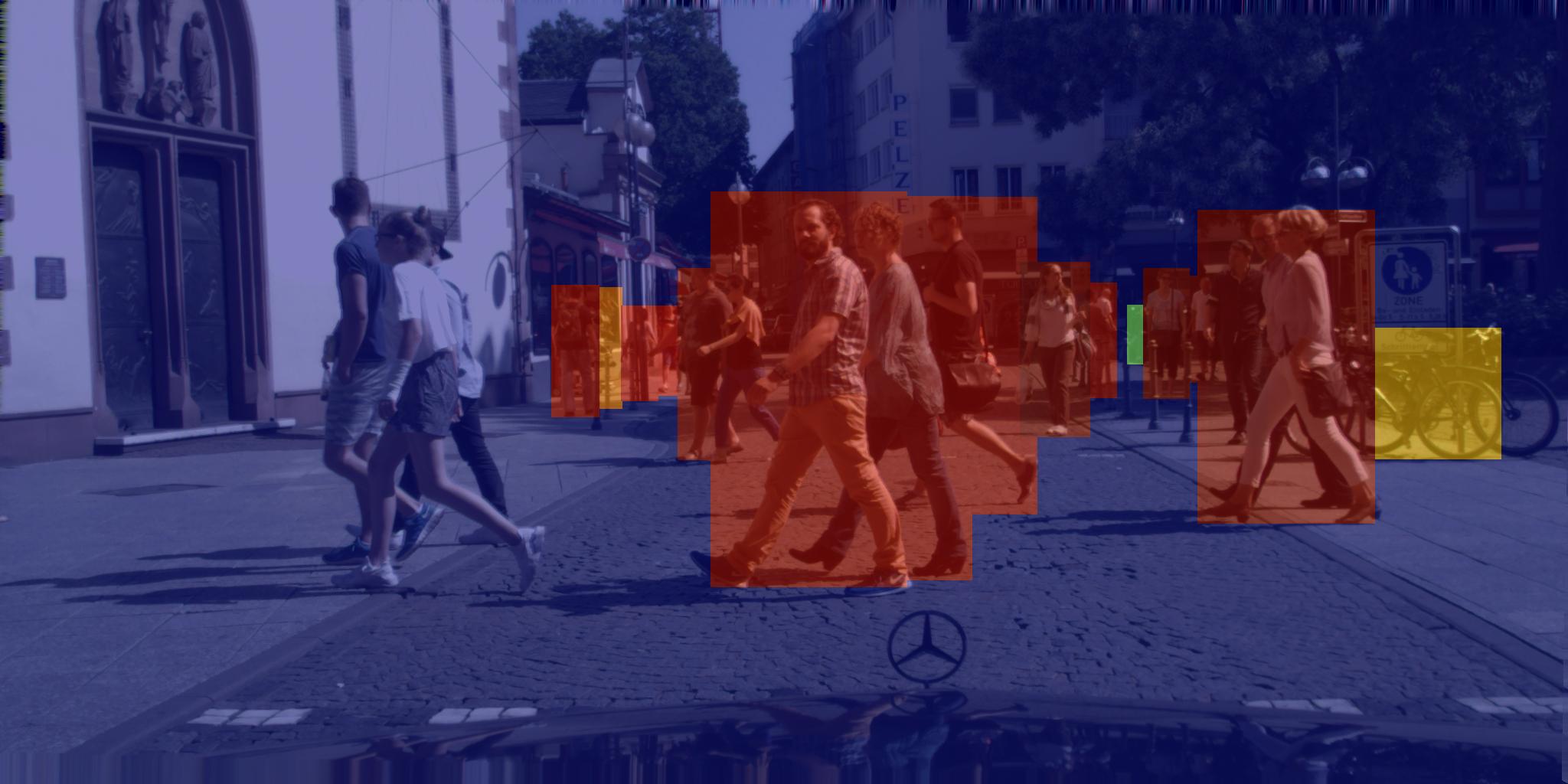}{}{}{}{Comparison of $obj$ (Top) vs. our $occ$ prediction (Bottom) on the Cityscapes dataset. The $occ$ score detects several objects that are not captured by the $obj$ score.}
\picSix{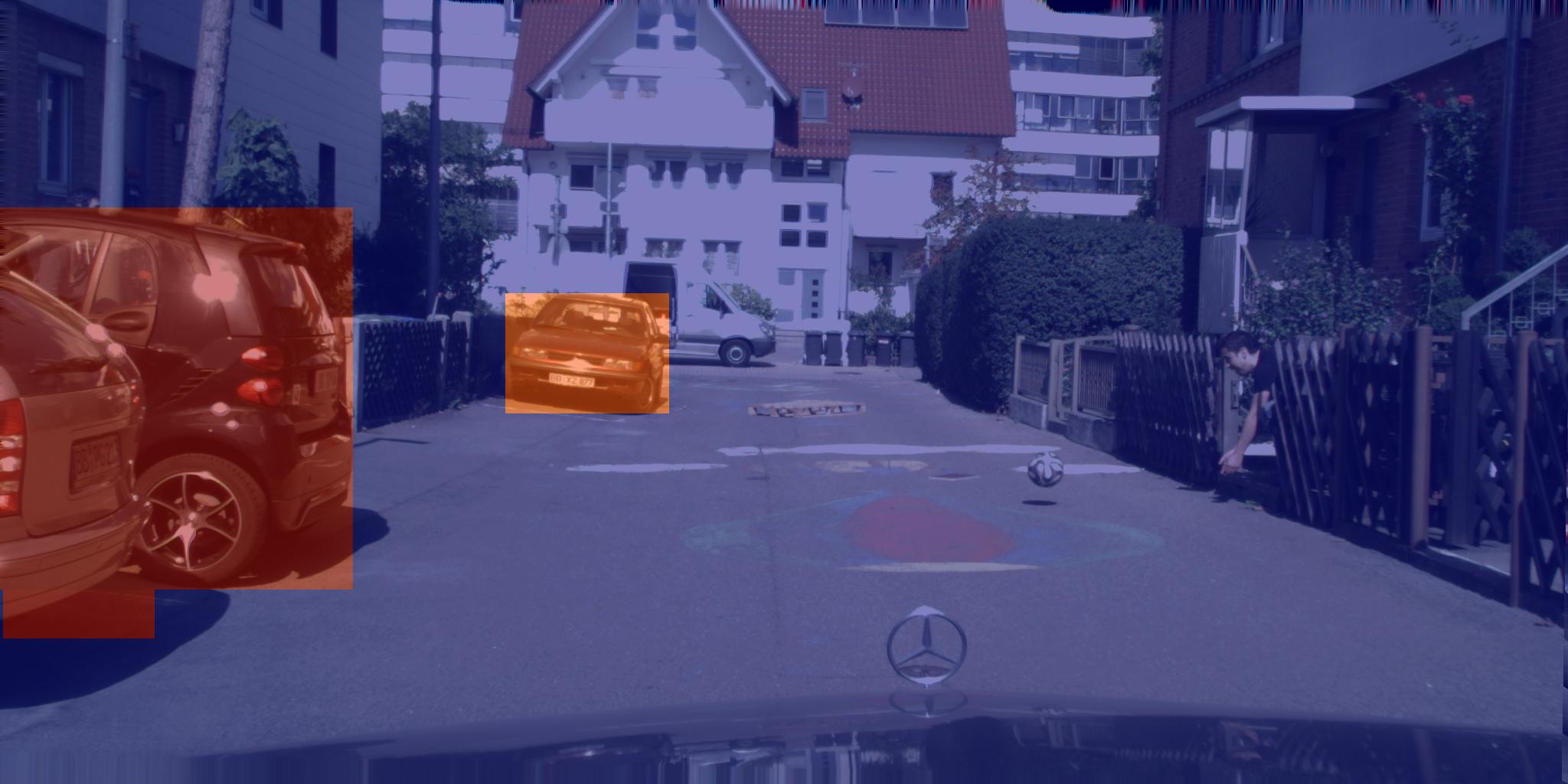}{}{}{}{Comparison of $obj$ (Top) vs. our $occ$ prediction (Bottom) on Fishyscapes dataset}
\picSix{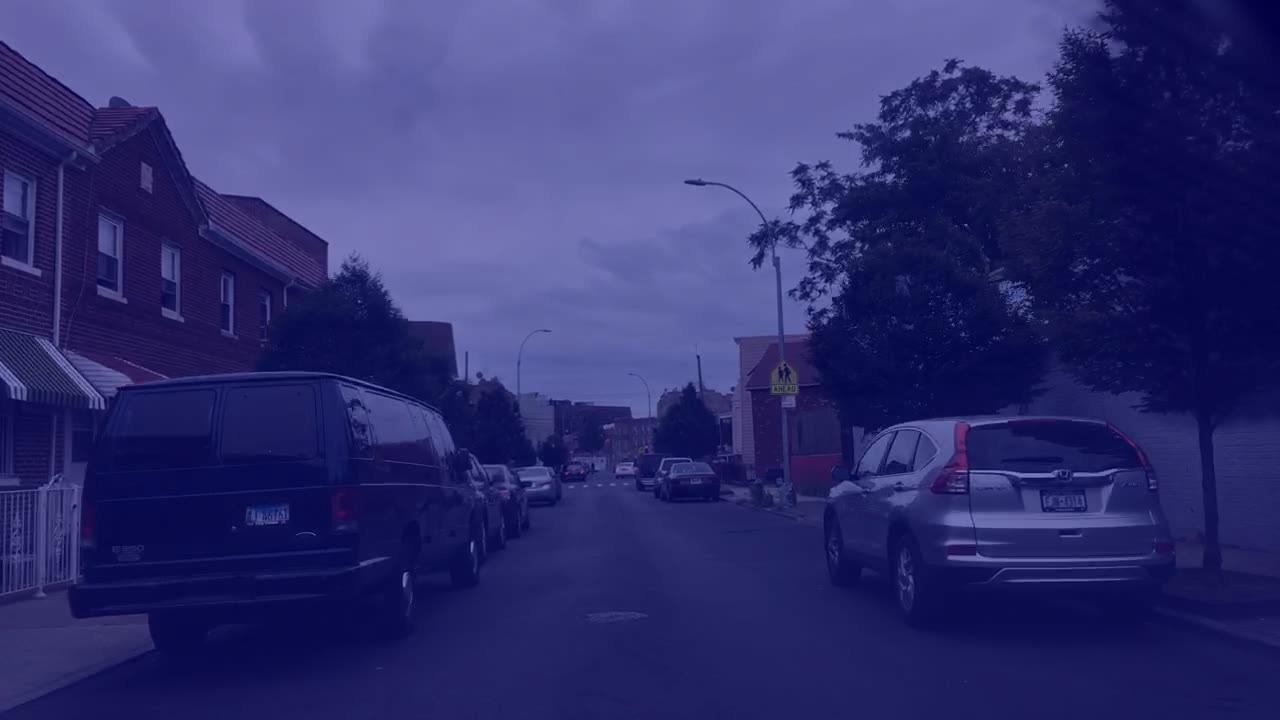}{}{}{}{Comparison of $obj$ (Top) vs. our $occ$ prediction (Bottom) on BDD100K dataset}
\picSix{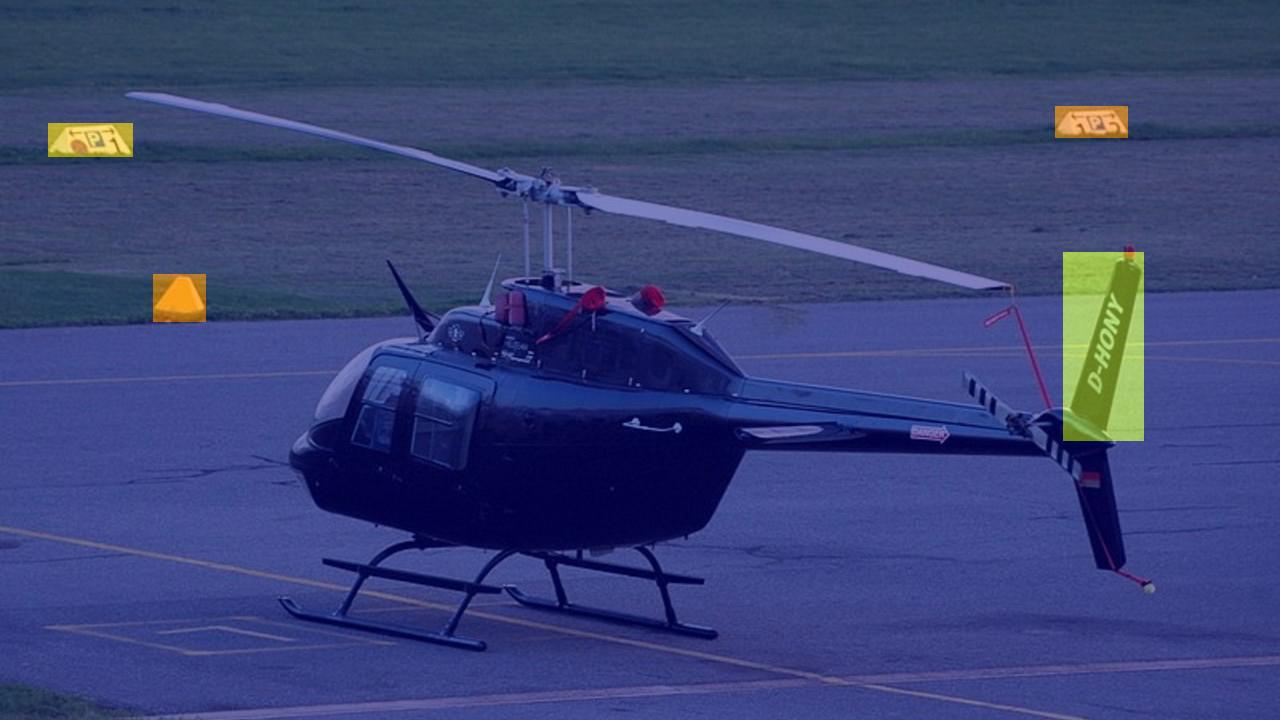}{}{}{}{Comparison of $obj$ (Top) vs. our $occ$ prediction (Bottom) on Anomaly dataset}
\picSix{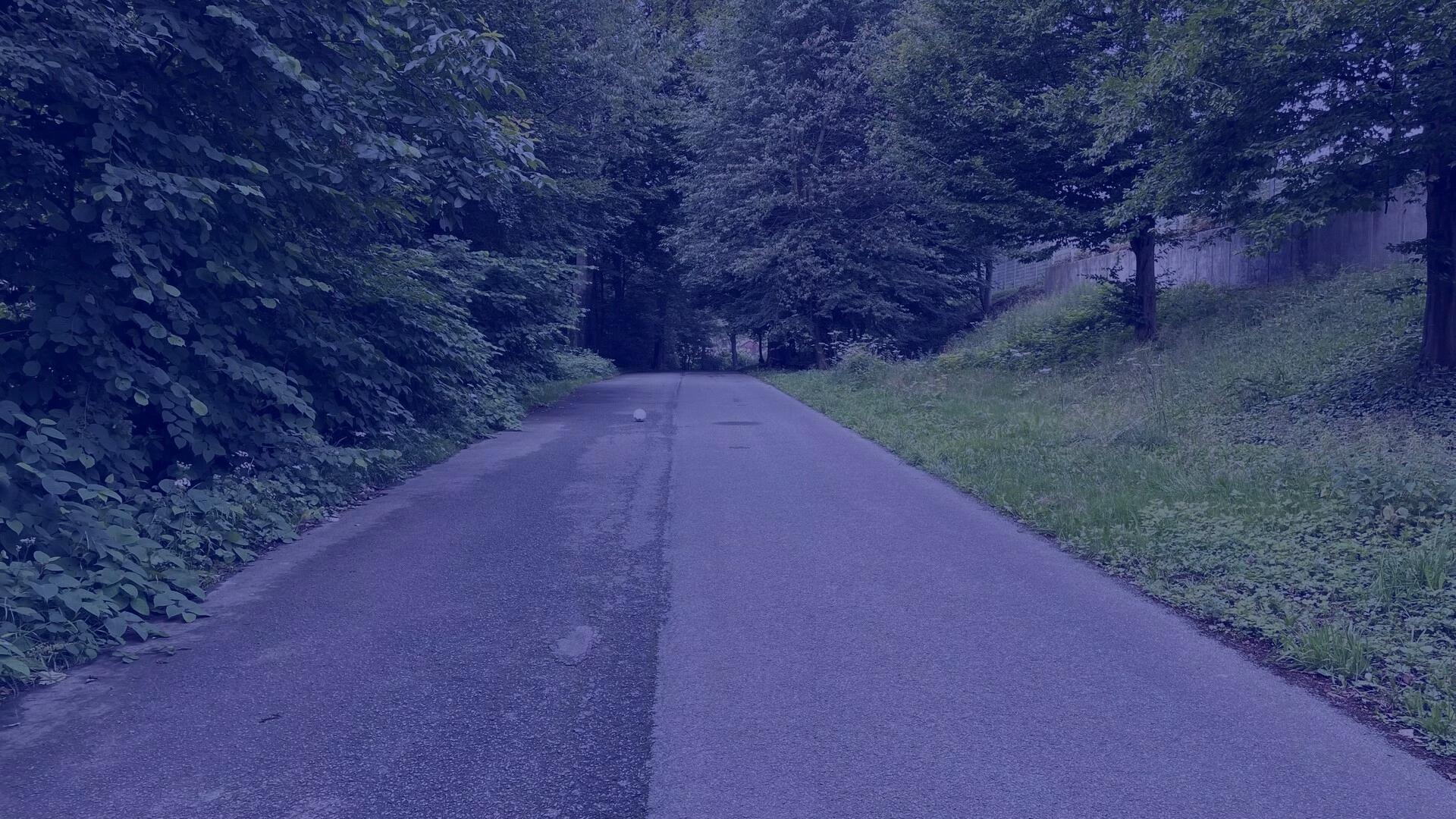}{}{}{}{Comparison of $obj$ (Top) vs. our $occ$ prediction (Bottom) on Obstacle dataset}

\FloatBarrier

\subsection{Reproducibility of Figure 7}
\label{supp:reproduce}

The full results for the presented \autoref{fig:dbse_table} are given in \autoref{tbl:DepthCS} and \autoref{tbl:DepthBDD}. 
\tblDepthsCS
\tblDepthsBDD

\end{document}